%% file: main.tex
\documentclass[11pt,a4paper]{article}
\usepackage[T1]{fontenc}
\usepackage[utf8]{inputenc}
\usepackage[margin=2.5cm]{geometry}
\usepackage{amsmath,amssymb}
\usepackage{array}
\usepackage{booktabs}
\usepackage{float}
\usepackage{graphicx}
\usepackage{makecell}
\usepackage{xcolor}
\usepackage{xurl}
\usepackage[super,sort&compress]{natbib}
\usepackage[section]{placeins}
\usepackage{caption}
\usepackage[hidelinks]{hyperref}

\newcolumntype{L}{l}
\newcolumntype{R}{r}
\newcommand{\tblwidth}{\linewidth}
\newcommand{\centeredtablecell}[1]{\begin{tabular}[c]{@{}c@{}}#1\end{tabular}}
\newcommand{\sitableformat}{\centering\sffamily\small}

\begin{document}
\title{Systematic Evaluation of TabPFN-TS and Chronos-2 for Zero-Shot Heat Load Forecasting in District Heating Networks}

\author{%
Ben Spoek\textsuperscript{1,*},
Karim K. Ben Hicham\textsuperscript{2},
Kai Derzsi\textsuperscript{1},\\
Philipp Althaus\textsuperscript{1},
Alexander Mitsos\textsuperscript{2},
and Dirk M\"uller\textsuperscript{1}\\[0.6em]
\small \textsuperscript{1}Chair of Energy Efficient Buildings and Indoor Climate, RWTH Aachen University,\\[-0.2em]
\small Mathieustr.\ 10, 52074 Aachen, Germany\\
\small \textsuperscript{2}Process Systems Engineering, RWTH Aachen University,\\[-0.2em]
\small Forckenbeckstr.\ 51, 52074 Aachen, Germany\\[0.4em]
\small \textsuperscript{*}Corresponding author: \href{mailto:ben.spoek@eonerc.rwth-aachen.de}{ben.spoek@eonerc.rwth-aachen.de}}

\date{}
\maketitle

\begin{abstract}
District heating energy hubs require reliable heat load forecasts for efficient operational scheduling. Forecasting models trained on historical data may require retraining as networks evolve. Zero-shot time-series foundation models and in-context forecasting therefore offer a promising alternative: they can adapt at inference time from recent observations rather than by repeated retraining. This study systematically evaluates TabPFN-TS and Chronos-2 for probabilistic heat load forecasting in two German district heating networks and compares them with trained baselines. Of particular interest is whether TabPFN-TS, whose underlying model is pretrained entirely on synthetic tabular rather than time-series data, can capture complex district heating dynamics. We analyze covariate choice, context length, temporal resolution, and forecast horizon on selected operating weeks, evaluate the selected configuration over the full year, and assess cross-network transfer. The principal benchmark assumes perfect weather forecasts; a separate sensitivity analysis uses retrospective weather predictions. Hourly 24-hour forecasting with a 12-week rolling context and ambient temperature provides a parsimonious configuration; longer context windows do not improve accuracy. Both TSFMs outperform all trained baselines in deterministic accuracy in the full-year benchmarks. Chronos-2 achieves the best deterministic scores, with TabPFN-TS remaining close: their CVRMSE values on the main data set are 12.48\% and 13.07\%, respectively. Chronos-2 also achieves lower continuous ranked probability scores in both networks, with TabPFN-TS remaining close. Finally, a TSFM-based Multi-Resolution Residual-Correction Forecaster combines an hourly base forecast with short-term high-resolution corrections. Relative to direct high-resolution forecasting, it generally reduces errors in total heat demand over 12-hour periods and recorded prediction times.
\end{abstract}


\section{Introduction}\label{sec:introduction}

To meet the climate targets of the European Union \cite{eu-law-climate}, the heat supply of buildings must be decarbonized. District heating networks (DHN) provide one promising pathway toward this goal \cite{lund4thGeneration2014} because they enable the integration of waste heat and renewable heat sources \cite{jodeiriRoleSustainableHeat2022}. Realizing this potential requires coordinating heat sources and storage under fluctuating availability, making accurate heat-demand forecasts particularly important for cost- and emission-optimal operation \cite{vandermeulenControllingDistrictHeating2018d}. Heat demand in DHNs is shaped by multiple superimposed drivers \cite{leiriaMethodologyEstimateSpace2023b}, including weather-dependent space-heating demand \cite{Wojdyga2008WeatherHeatDemand}, comparatively stochastic domestic hot water consumption \cite{ivankoSplittingMeasurementsTotal2021a}, and additional operational, daily, and seasonal variations \cite{dangDailySeasonalHeat2022}.

DHNs evolve as buildings are connected or disconnected, renovated, equipped with additional technologies, or subject to changes in occupancy and usage. These changes may not be fully known to the operator and can alter the underlying heat-demand dynamics.\cite{derzsiConceptDriftDetection2026} Data-driven forecasting methods can leverage operational observations, but conventional supervised machine-learning models typically require explicit training and retraining in response to such changes.\cite{celikAdaptationStrategiesAutomated2021} We hypothesize that in-context learning (ICL), which conditions predictions on supplied observations without updating model weights, can support adaptation to concept drift as a sliding context window continually incorporates recent observations. Whether such models can rapidly distinguish relevant post-change observations from older context remains an open question.

Recent advances in foundation models, models pretrained for reuse across different downstream tasks, are highly promising in this regard. TabPFN~\cite{hollmann2025accurate} and TabICL~\cite{qu2025tabicl} achieve strong performance on tabular prediction benchmarks through ICL without task-specific training. The time-series extension TabPFN-TS~\cite{hoo2025tabpfnts} applies time-series featurization to TabPFN and is competitive with leading forecasting models~\cite{obermeier2026fets}. Other foundation models, such as Chronos-2~\cite{ansariChronos2UnivariateUniversal2025}, are pretrained directly for time-series forecasting. Both approaches enable zero-shot forecasting from historical observations without task-specific weight training. However, it remains unclear whether TabPFN-TS's synthetic tabular prior adequately captures the complex heat-demand dynamics of DHNs, or whether time-series-specific pretraining provides an advantage. We therefore systematically evaluate these two contrasting time-series foundation model (TSFM) approaches to probabilistic heat load forecasting. We compare their accuracy, predictive uncertainty and operational requirements, using classical statistical, machine-learning and deep-learning models as trained benchmarks.

The contributions of this work are as follows:

\begin{itemize}
\item We provide a systematic evaluation of two TSFMs, TabPFN-TS and Chronos-2, for probabilistic heat load forecasting in DHNs.
\item We investigate how covariate choice and context length affect TSFM forecasts to identify configurations that achieve strong performance with limited input-data requirements.
\item We compare hourly and 15-minute resolutions to assess whether TSFMs can capture higher-frequency heat load dynamics that are relevant for district heating operation.
\item We evaluate forecast horizons from 4 hours to one week to characterize horizon-dependent TSFM performance and its interaction with suitable context length.
\item We compare both foundation models with trained forecasting baselines and evaluate cross-network transfer.
\item We derive and evaluate a Multi-Resolution Residual-Correction (MRRC) Forecaster based on the diagnostic findings.
\end{itemize}

\section{Related Work}\label{sec:related_work}

This section reviews established forecasting approaches for DHNs, the foundation-model concepts underlying TabPFN-TS and Chronos-2, recent energy-forecasting benchmarks, and residual-correction architectures relevant to the present study.

\subsection{Heat Load Forecasting in District Heating Networks}\label{subsec:heat_load_forecasting}

Heat load forecasting methods range from physically informed models to statistical, machine-learning, and deep-learning approaches.\cite{ntakoliaMachineLearningApplied2022} Physically motivated and gray-box methods incorporate knowledge of building and district-heating dynamics \cite{Dotzauer2002SimpleLoadPrediction,Nielsen2006GreyBoxHeatConsumption}, while engineered load-shape indicators provide a more empirical representation of operational demand patterns \cite{gaddDailyHeatLoad2013}. Classical data-driven approaches formulate forecasting as supervised regression, including linear regression and autoregressive models with exogenous inputs \cite{fangEvaluationMultipleLinear2016,Dahl2017EnsembleWeatherPredictionsDH}.

Machine-learning methods extend these formulations through nonlinear relationships. Common examples include support vector regression \cite{Park2010HeatConsumptionPLSANN_SVR}, decision-tree models \cite{Finkenrath2022HolisticThermalLoadDispatch}, and gradient-boosting methods such as XGBoost \cite{Runge2023AIModelsDistrictHeatingDemand}. Neural approaches include feed-forward networks \cite{Izadyar2015ResidentialHeatingMonthlyGas}, nonlinear autoregressive models with exogenous inputs \cite{Powell2014DistrictEnergyLoadForecasting}, convolutional-recurrent architectures \cite{Song2021HybridCNNLSTMHeatingLoad}, recurrent neural networks \cite{Kato2008RNNHeatDemandDHC}, attention-based long short-term memory (LSTM) networks \cite{Xue2020AttentionLSTMHeatLoadXingtai}, and Transformer-based models such as Informer \cite{Gong2022InformerDHLoad}. Rather than replacing earlier methods, these model families provide increasingly flexible alternatives for representing heat-demand dynamics.

Most DHN load-forecasting models are fitted to a specific network and may require retraining as system dynamics change.\cite{gamaSurveyConceptDrift2014} Adapting to concept drift remains challenging \cite{luLearningConceptDrift2018}, as changes must be detected, suitable adaptation windows selected, and recurring concepts accounted for \cite{webbCharacterizingConceptDrift2016}. Such recurrence is particularly relevant in district-heating systems, where seasonal shifts in the relative dominance of space-heating and domestic-hot-water demand create recurring regimes \cite{leiriaMethodologyEstimateSpace2023b}. These limitations motivate forecasting methods that can adapt through context rather than repeated parameter fitting. The following subsection introduces in-context learning for tabular prediction and its extension to time-series forecasting.

\subsection{Tabular Prediction with TabPFN and Its Extension to Time-Series Data}\label{subsec:tabular_pfn}

TabPFN~\cite{hollmann2025accurate} is a foundation model for tabular prediction based on in-context learning. It is pretrained on a large collection of fully synthetic supervised-learning tasks and, at inference time, predicts query labels from labeled context examples.
Unlike conventionally trained neural networks and gradient-boosted models, TabPFN requires no task-specific parameter fitting.\cite{hollmann2023tabpfn,hollmann2025accurate, qu2025tabicl}

More formally, the model receives both labeled context examples \(D_c=(X_c,y_c)\) and unlabeled query examples \(x_*\) and produces predictions \(y_*\) in a single forward pass. In this sense, the training data of the target task become context data for the pretrained model. The Bayesian interpretation is central to Prior-Data Fitted Networks (PFNs).\cite{muller2022transformers} Let \(\phi\) denote a latent data-generating task or process. The posterior predictive distribution can be written as
\begin{equation}
p(y_* \mid x_*,D_c) = \int p(y_* \mid x_*,\phi)\,p(\phi \mid D_c)\,d\phi .
\end{equation}
Using Bayes' rule,
\begin{equation}
p(\phi \mid D_c) = \frac{p(D_c \mid \phi)p(\phi)}{p(D_c)},
\end{equation}
so that
\begin{equation}
\begin{aligned}
p(y_* \mid x_*,D_c)
&= \dfrac{1}{p(D_c)}
\int p(y_* \mid x_*,\phi)\,
       p(D_c \mid \phi)\,p(\phi)\,d\phi \\[0.5em]
&\propto
\int p(y_* \mid x_*,\phi)\,
       p(D_c \mid \phi)\,p(\phi)\,d\phi .
\end{aligned}
\end{equation}
Computing this integral explicitly would generally be intractable. A PFN instead amortizes the computation: it is trained to map \((D_c,x_*)\) directly to an approximation of the posterior predictive distribution in a forward pass. This approximation is Bayesian with respect to the synthetic pretraining prior over data-generating tasks, rather than an explicit posterior over the real-world target system. The practical consequence is that TabPFN removes the standard per-data-set gradient-based training loop in its default use.\cite{hollmann2023tabpfn,hollmann2025accurate} No classical hyperparameter search is required for basic application, although inference-time ensembling, preprocessing variants, fine-tuning, or hyperparameter tuning can still be used to improve performance.

TabPFN-TS, as proposed by Hoo et al. \cite{hoo2025tabpfnts}, can be directly used for time series forecasting by reformulating a time series as a tabular regression problem. Each time step becomes a row in a table. The feature set includes temporal features such as a running time index, cyclic calendar features, a non-cyclic year feature, automatically extracted seasonal features, and optionally time-varying covariates whose future values are known at prediction time. The historical observations form the labeled context \(D_c\), while the future time points form query rows with known features \(x_*\) and unknown targets \(y_*\). TabPFN then outputs a probabilistic forecast for the full prediction horizon in a non-autoregressive forward pass. This is notable because the same study reports strong GIFT-Eval~\cite{aksu2024gift} performance for TabPFN-TS, although the underlying TabPFN model was pretrained only on synthetic non-temporal tabular data. The synthetic pretraining paradigm is important for the present application: since TabPFN is pretrained on synthetic supervised prediction tasks rather than on collections of real district-heating time series, the risk of direct or indirect pretraining-test overlap with historical district-heating backtesting data is eliminated. In addition, the probabilistic forecast can be interpreted as an approximation to the posterior predictive distribution under the learned synthetic prior. This makes TabPFN-TS particularly interesting for heat load forecasting, where reliable uncertainty estimates are operationally relevant and where point forecasts alone do not fully describe the stochastic demand process. Hoo et al. \cite{hoo2025tabpfnts} report that TabPFN-TS can distinguish signal from noise, adapt its probabilistic predictions accordingly, incorporate covariates, and handle seasonal patterns, while trend recognition, especially for linear and exponential trends, remains a reported weakness.

\subsection{Time-Series Forecasting with Chronos-2}\label{subsec:chronos2}

Chronos-2, proposed by Ansari et al. \cite{ansariChronos2UnivariateUniversal2025}, is a time-series foundation model capable of probabilistic predictions. The model uses normalization of the covariates and targets. Additionally, it applies patching where each patch represents several consecutive observations. Its transformer architecture combines temporal attention and group attention, which allows relationships across time and across targets and covariates to be captured. The model supports both past-only covariates and covariates with known future values. Chronos-2 is pretrained on time-series forecasting tasks using real and synthetic data. It differs from TabPFN-TS in its architecture and pretraining assumptions.

Since Chronos-2 was partially pretrained on real-world time-series data, potential train-test contamination between the pretraining corpus and the downstream evaluation data must be considered. Based on the disclosed Chronos-2 pretraining corpus, leakage of time-series training data into the 2024 heat load forecasting data sets for the DHNs in Munich or Flensburg appears unlikely. The real pretraining data sets listed in Appendix A of the Chronos-2 documentation are univariate time series from broad domains such as electricity, solar and wind generation, weather, transport, web, cloud operations, and simulated U.S. building energy, but no district-heating heat load data from Munich or Flensburg are reported \cite{ansariChronos2UnivariateUniversal2025}. Moreover, Chronos-2's multivariate and covariate-informed training tasks are stated to rely entirely on synthetic data.

\subsection{Foundation-Model Benchmarks in Energy Forecasting}\label{subsec:tabpfn_ts_energy_benchmarks}

Recent work indicates that time-series foundation models can be useful in energy-forecasting applications. Meyer et al. \cite{meyer2025householdtsfm} show that foundation models can provide performance comparable to approaches trained from scratch for household electricity-demand forecasts. Koch et al. \cite{koch2026thermalgemschronos2} study thermal-energy forecasting and compare multisource transfer-learning strategies with time-series foundation models, including TimesFM, Toto, and Chronos-2. Chronos-2 achieves the strongest performance among the considered time-series foundation models. They further find that multisource transfer learning requires at least 16 building data sets before it outperforms time-series foundation models.

Comparative benchmarks including both TabPFN-TS and Chronos-2 in energy forecasting provide the closest context for this study: Hertel et al. \cite{hertel2026explainable} compare TabPFN-TS and Chronos-2 for load forecasting in electricity grids at transmission-system-operator level and use Shapley-based explanations to increase interpretability. In their study, Chronos-2 outperforms TabPFN-TS in both the univariate and covariate settings. Obermeier et al. \cite{obermeier2026fets} benchmark several time-series foundation models on energy-forecasting tasks, including Chronos-2 and TabPFN-TS in a covariate setting. The benchmark includes a data set from Flensburg's DHN as part of a heat load prediction subset. For this heat load subset and overall, covariate models outperform univariate models. Chronos-2 again outperforms TabPFN-TS overall and on the heat load subset, although TabPFN-TS obtains some wins over Chronos-2 in the covariate setting on other subcategories and several wins overall. These findings motivate evaluating both approaches under the same district-heating forecasting conditions.

TabPFN has also been successfully applied to energy-related regression tasks. Liu et al. \cite{liu2026tabpfnsolar} evaluate TabPFN for regression tasks in solar-energy meteorology and report that TabPFN outperforms multiple regression, artificial neural networks, Random Forest, and Gradient Boosting. However, they do not directly address TabPFN-TS as a time-series forecasting method.

Alkhulaifi et al. \cite{alkhulaifiAutoEnergyAutomatedFeature2025} show the importance of feature engineering for time-series forecasting with TabPFN. TabPFN-TS largely automates this step by embedding generic temporal and seasonal feature construction into the forecasting pipeline.

Ramachandran et al. \cite{ramachandran2026heatdemandchronos2} compare several time-series foundation models and classical machine-learning baselines with a method that combines a continuous wavelet transform with LSTMs for heat-demand forecasting. Their proposed method outperforms the benchmark approaches. However, it is specifically engineered for the considered forecasting problem and is therefore less directly transferable than general-purpose time-series foundation model approaches.

Overall, the literature motivates a systematic comparison of TabPFN-TS and Chronos-2 for zero-shot probabilistic heat load forecasting. Open questions concern their application-specific configurations, deterministic and probabilistic performance, computational requirements, comparison with trained baselines, and transfer across DHNs with different characteristics. Their different pretraining approaches additionally allow us to examine how competitive a synthetic tabular prior remains against a model pretrained directly for time-series forecasting.

\subsection{Residual-Correction Architectures for Time-Series Foundation Models}\label{subsec:related_residual_correction}

Residual correction of TSFM forecasts and the use of TSFMs as residual forecasters are emerging directions in time-series forecasting. Chitsaz and Haratizadeh \cite{chitsazBridgingUnivariateFoundation2025} investigate TSFM-based residual correction for several regression and machine-learning models in stock-price forecasting. They compare this approach with a relational architecture in which a TSFM separately forecasts the stock price and auxiliary financial variables, and the concatenated forecasts are passed to a multivariate predictor. Residual correction reduces mean squared error (MSE) relative to the standalone multivariate baseline for all five model classes and outperforms the relational architecture in four of five comparisons. For price-direction classification, the relational architecture achieves a higher macro-averaged F1-score in three comparisons, residual correction in one, with one tie.

In contrast, Dai et al. \cite{daiLearningContextErrors2026} use a frozen TSFM to generate the base forecast. Their Online Residual Contextual Adaptation (ORCA) framework estimates an additive residual correction using a linear adapter that decomposes the concatenated historical input and base forecast into trend and seasonal components. The base and corrected forecasts are then combined using a Boltzmann softmax weighting mechanism based on recent prediction errors. Across eight data sets covering energy, finance, meteorology, and traffic, the authors report an average MSE reduction on every data set, with results aggregated over five TSFMs and three forecasting horizons.

Gao et al. \cite{gaoLightweightWrappersAdapting2026} propose $\mathrm{SMR}^{2}$, a TSFM wrapper that constructs forecasts using the same frozen TSFM with context data at different resolutions. A separate ridge-regression model predicts a residual correction for each resolution-specific forecast. The corrected forecasts are then combined using weights determined from validation performance. The authors report consistent improvements across several TSFMs for regional drought forecasting.

Related residual-correction schemes have also been investigated for Transformer-based forecasting models. Biswas et al. \cite{biswasOneStepCloser2026} train a base model on the original time series and a second model to forecast its residuals. The predicted residuals are scaled before being used to correct the base forecast. Both models use the same look-back length and prediction horizon. Their main iTransformer-based experiments show improved accuracy over the uncorrected model across all eight data sets from energy, transportation, meteorology, and finance.

These studies motivate adapting residual-correction architectures to the distinct heat-load dynamics and operational requirements of DHNs. In particular, they leave open how to combine longer-horizon planning forecasts with frequently updated, high-resolution residual corrections. Within the reviewed literature, using a TSFM to forecast the residuals of a TSFM prediction remains underexplored. The present study investigates this combination as an application-specific adaptation of residual correction.

\section{Methodology}\label{sec:methodology}

This section describes the forecasting models and rolling prediction setup, followed by the procedures for evaluating deterministic accuracy, probabilistic forecast quality and computational cost. It then defines the anomaly-screening criteria and the metric used to quantify interannual distributional change, before introducing the MRRC forecasting architecture.

\subsection{Forecasting Models and Prediction Setup}\label{subsec:forecasting_setup}

The forecasting workflow uses rolling historical context windows comprising past heat load and observed weather variables. Known future covariates comprise weather variables over the prediction horizon and calendar features. Unless stated otherwise, ambient temperature is the only weather variable. The prediction target is the future heat load. Both foundation models are evaluated without task-specific weight training and produce quantile forecasts. The median is used as the deterministic prediction, while the stored quantiles support probabilistic evaluation.

TabPFN-TS and Chronos-2 are evaluated on the same operational forecasting task to compare their different pretraining paradigms and inference architectures. Both models receive the same future covariate values over the prediction horizon and, where permitted by model constraints, the same rolling context windows. The maximum number of context points that can be passed to Chronos-2 is 8192 time steps. If a selected context window exceeds this limit, only the most recent 8192 time steps are passed as context to Chronos-2, and results are reported at the effective context length.

AutoGluon TimeSeries \cite{celikAdaptationStrategiesAutomated2021, pmlr-v224-shchur23a} is used as a benchmark suite for the selected hourly day-ahead forecasting task. The benchmark contains several model families. Persistence baselines are represented by Naive and Seasonal naive. Naive repeats the last observed heat load value over the prediction horizon. Seasonal naive uses a seasonal period of 24 hours and therefore predicts each future hour from the corresponding observed value on the previous day. Classical statistical baselines include exponential smoothing (ETS), automatic ETS (AutoETS), and automatic autoregressive integrated moving average (AutoARIMA). Tabular machine-learning baselines include direct LightGBM (LGBM) and XGBoost models as well as a recursive LGBM model. Neural baselines include DeepAR, TemporalFusionTransformer (TFT), Feed-forward, and DLinear. AutoGluon's weighted ensemble is reported as AutoGluon Ensemble.

The predictors are configured with ambient temperature as a known covariate, so future ambient-temperature values are supplied over each prediction horizon. Models with native known-covariate support can use this information directly, including the direct and recursive tabular models, DeepAR, and TemporalFusionTransformer. Other models in the benchmark, such as the persistence, statistical, Feed-forward, and DLinear baselines, do not use the future covariate in the same way.

These AutoGluon benchmarks fitted once are not constrained to the same rolling context used for the selected foundation-model setup. Instead, each prediction call receives the observed target history from the start of 2023 up to the respective forecast start. Individual AutoGluon model classes then use this history according to their own defaults: statistical models may truncate long series internally, neural models use model-specific context lengths, and tabular models construct lag, time, and covariate features.

In the main experiments, future ambient temperature is provided as the realized temperature over the prediction horizon. This setting can be interpreted as a perfect-weather-forecast assumption: the heat load forecasting problem is isolated from errors in numerical weather prediction, and model behavior is evaluated conditional on known future weather.

All experiments were run on the RWTH Aachen University HPC cluster. Each job used one NVIDIA H100 GPU with 95,830 MiB memory, 8 CPU cores, and 64 GB RAM. Nodes are equipped with Intel Xeon Platinum 8468 CPUs, with two sockets and 48 physical cores per socket. The software environment used Python 3.12.3 and CUDA 12.6.3, with tabpfn-time-series 1.1.0, tabpfn 8.0.3, chronos-forecasting 2.2.2, and autogluon.timeseries 1.5.0.

\subsection{Evaluation Metrics and Approach for Model Comparison}\label{sec:evaluation}

For experiments with overlapping forecast windows, metrics are computed after collapsing predictions to a single operational trajectory: for each target timestamp, the forecast with the latest forecast start at or before the target timestamp is retained. Thus each realization contributes once to the error metrics.

Point-forecast quality is assessed using the coefficient of variation of the root mean squared error (CVRMSE), the coefficient of determination (\(R^2\)), and mean absolute error (MAE). For \(n\) evaluation samples, let \(y_t\) denote the target value, \(\hat{y}_t\) the corresponding point forecast, and \(\bar{y}=\frac{1}{n}\sum_{t=1}^{n}y_t\) the mean target value of the evaluation set. In this study, \(y_t\) corresponds to the measured heat load.

The three metrics are defined as
\begin{align}
\mathrm{CVRMSE} &= 100 \cdot
\frac{
\sqrt{\frac{1}{n}\sum_{t=1}^{n}\left(\hat{y}_t-y_t\right)^2}
}{
\bar{y}
}, \\
R^2 &= 1 -
\frac{
\sum_{t=1}^{n}\left(y_t-\hat{y}_t\right)^2
}{
\sum_{t=1}^{n}\left(y_t-\bar{y}\right)^2
}, \\
\mathrm{MAE} &= \frac{1}{n}\sum_{t=1}^{n}\left|\hat{y}_t-y_t\right|.
\end{align}

CVRMSE normalizes the root mean squared error by the mean target value, making the error comparable across operating regimes and between the two DHNs.

For longer-horizon planning, we evaluate the total heat demand over rolling 12-hour windows starting at hourly intervals. Let \(E_i\) and \(\hat E_i\) denote the observed and predicted integrated heat demand, respectively, using the forecast available at anchor \(i\), and let \(\bar E\) be the mean observed integrated heat demand across the \(N\) evaluated windows. Energy CVRMSE (E-CVRMSE) and energy bias (E-bias) are defined as
\begin{align}
\mathrm{E\text{-}CVRMSE}
&= \frac{100}{\bar E}
\sqrt{\frac{1}{N}\sum_{i=1}^{N}(\hat E_i-E_i)^2},
\label{eq:energy_cvrmse}\\
\mathrm{E\text{-}bias}
&= \frac{100}{N\bar E}
\sum_{i=1}^{N}(\hat E_i-E_i).
\label{eq:energy_bias}
\end{align}
Both metrics are expressed as percentages; negative E-bias indicates underprediction and positive E-bias indicates overprediction.

Probabilistic forecast quality is assessed from the stored quantile forecasts using the continuous ranked probability score (CRPS). Let \(F_t\) denote the predictive cumulative distribution function for time step \(t\), where \(y_t\) is the observed heat load, and let \(\mathbf{1}\{\cdot\}\) denote the indicator function, which equals one if the condition in braces is true and zero otherwise. The CRPS is defined as

\begin{equation}
\mathrm{CRPS} =
\frac{1}{n}\sum_{t=1}^{n}
\int_{-\infty}^{\infty}
\left(F_t(z)-\mathbf{1}\{z\ge y_t\}\right)^2\,dz.
\end{equation}

Since the probabilistic forecasts are stored as discrete quantiles, CRPS is approximated by twice the trapezoidal integral of the pinball loss over the stored quantile levels from 0.01 to 0.99, averaged over the evaluated time steps. Lower CRPS values indicate better overall probabilistic forecast quality.

Central prediction intervals are additionally described by their absolute and relative widths. Relative interval width is the mean q90--q10 width divided by the mean observed heat load, expressed as a percentage.

Calibration is assessed using the probability integral transform (PIT), defined as
\begin{equation}
\mathrm{PIT}_t = F_t(y_t).
\label{eq:pit}
\end{equation}
PIT values are approximated from the stored quantile forecasts. For calibrated continuous predictive distributions, PIT values are uniformly distributed. The histograms pool forecast horizons and dates.

For the full-year model comparison, daily forecast blocks are also used to compare the consistency of the algorithms over time. Each issued 24-hour forecast is treated as one block, and CVRMSE, \(R^2\), and MAE are computed over the 24 hourly prediction points for each model and forecast start. Models are ranked within each block, using ascending ranks for CVRMSE and MAE and descending ranks for \(R^2\); ties are assigned average ranks. The reported mean ranks therefore describe how consistently a model performs well across individual forecast days, not only its aggregate full-year error. Pairwise win-rate matrices and critical difference (CD) diagrams use daily CVRMSE comparisons. The CD diagrams use a nominal significance level of 0.05.

For full-year evaluations, uncertainty in aggregate metrics is quantified using 95\% percentile confidence intervals from 10,000 paired circular block-bootstrap samples. Blocks comprise seven consecutive forecast days, retaining complete issued forecasts and using identical resampled blocks across models.

Computational cost is separated into model loading, estimated first-call overhead and steady-state inference. Steady-state inference time is calculated as the mean prediction-call duration from the second to the final call, excluding surrounding input preparation and prediction post-processing. Estimated first-call overhead is the first prediction-call duration minus the mean duration of subsequent calls and is reported separately from model loading. 

To quantify computational cost, the mean steady-state prediction-call time and the operational real-time factor (RTF) are reported. The operational real-time factor is defined as the ratio between the total recorded prediction-call time and the total elapsed operating time over the evaluation period \(T_{\mathrm{eval}}\),
\begin{equation}
\mathrm{RTF}_{\mathrm{op}} = \frac{\sum_i t_{\mathrm{comp},i}}{T_{\mathrm{eval}}},
\end{equation}
where \(t_{\mathrm{comp},i}\) denotes the computation time of forecast \(i\). This metric captures the computational burden of a forecasting scheme in an operational setting, accounting for both the cost of individual forecasts and their update frequency. Values below one indicate that the forecast is computed faster than real time. 

\subsection{Anomaly Screening and Exclusion}\label{subsec:anomaly_screening}

Potentially anomalous heat-load observations are identified using a model-independent screening proxy: nonpositive heat load or at least three consecutive exactly identical values at the data set's native resolution. These patterns can indicate maintenance, faults or signal outages, but do not establish their cause. As independently verified anomaly labels are unavailable, flagged observations are treated as suspected anomalies.

Weeks containing flagged observations are excluded from configuration selection to prevent brief anomalous periods from driving the configuration choices. The principal full-year benchmark retains the recorded heat-load series without anomaly-specific modification. This evaluates the selected configuration under continuous operational conditions and follows a data-driven setting in which no manual preprocessing of the target signal is applied for the final assessment.

\subsection{Quantification of Distributional Change}\label{subsec:distributional_change}

To characterize interannual changes in the observed heat-load distribution, we compute the normalized Hellinger distance between consecutive years.
\begin{equation}
H(p,q)=\sqrt{\frac{1}{2}\sum_b\left(\sqrt{p_b}-\sqrt{q_b}\right)^2},
\label{eq:hellinger_distance}
\end{equation}
where \(p_b\) and \(q_b\) are consecutive years’ histogram probabilities for each of the 470 bins indexed by \(b\). Histograms use all available raw 15-minute heat loads without anomaly exclusion and 470 shared equal-width bins determined on pooled 2020--2024 observations.

\subsection{Multi-Resolution Residual-Correction Forecaster}\label{subsec:mrrc_forecaster_method}

\begin{figure}[!htbp]
  \centering
  \includegraphics[width=\textwidth]{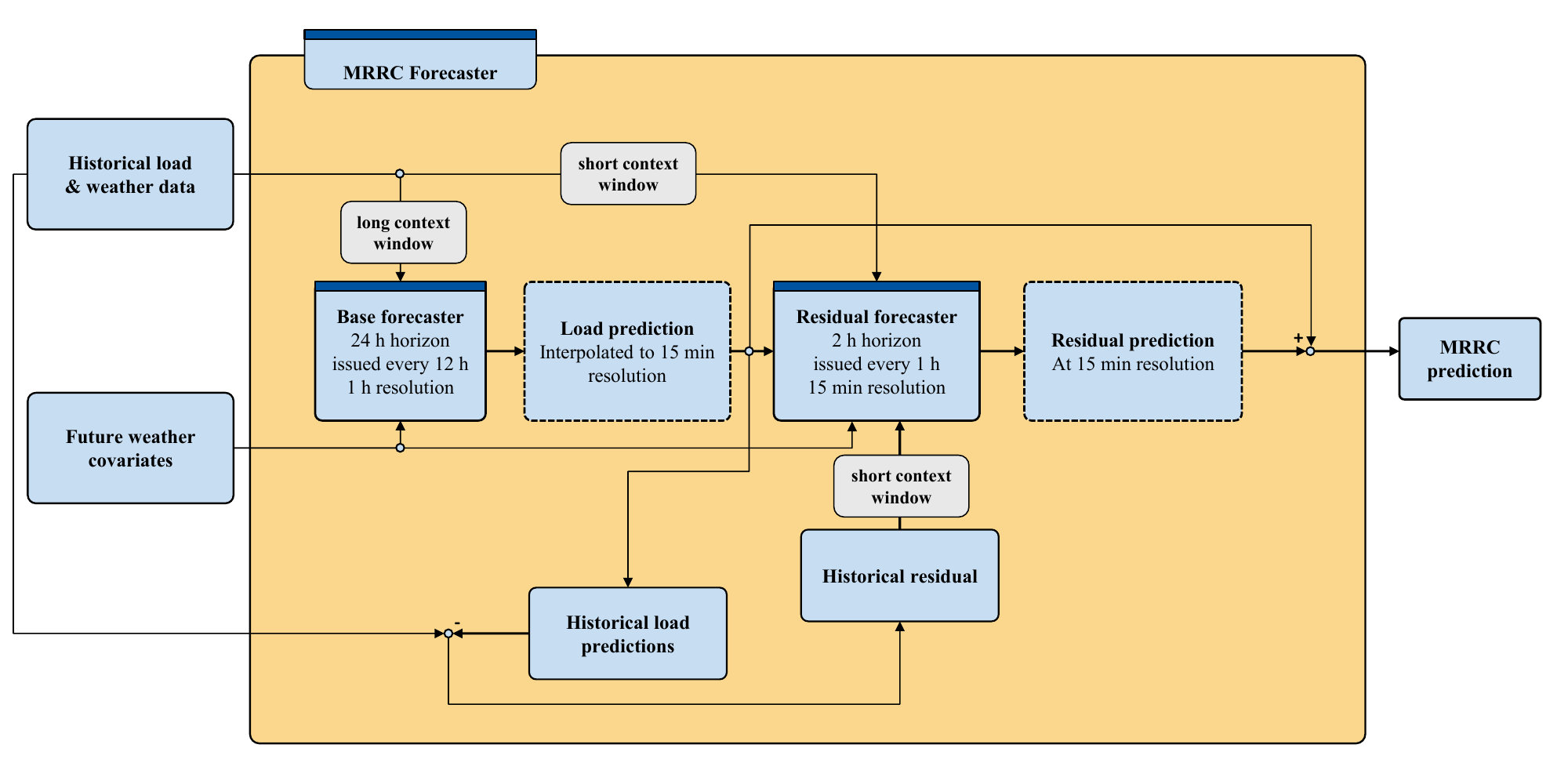}
  \caption{Schematic structure of the Multi-Resolution Residual-Correction Forecaster. }
  \label{fig:mrrc_forecaster_setup}
\end{figure}

Building on the residual-correction approaches reviewed in Section~\ref{subsec:related_residual_correction} and our configuration findings, we develop the MRRC Forecaster as a follow-up extension of the evaluated forecasting approaches. It combines a long-horizon base forecast with a short-horizon residual correction at higher temporal resolution. The architecture is illustrated in Figure~\ref{fig:mrrc_forecaster_setup}. 

The first component, referred to as the Base Forecaster, provides the general daily load trajectory potentially needed for scheduling and storage planning. Its point forecast is linearly interpolated to the higher temporal resolution of 15 minutes. The second component, referred to as the Residual Forecaster, forecasts the remaining high-frequency error of this baseline. The final point prediction is obtained by adding the predicted residual to the interpolated base forecast. The architecture aims to improve the representation of short-term dynamics while retaining a longer planning horizon. The MRRC Forecaster is evaluated here as a point-forecast architecture.

From a boosting perspective, the setup can be interpreted as a single residual-correction step. An initial model \(F(x)\) produces the base prediction, while a second model \(h(x)\) predicts the residuals \(r = y - F(x)\). The final prediction is then given by \(\hat{y} = F(x) + h(x)\). In the MRRC Forecaster, the Base Forecaster corresponds to \(F(x)\), and the Residual Forecaster corresponds to \(h(x)\).

Historical residual context is constructed by subtracting the interpolated base forecasts available at each historical timestamp from the observed high-resolution heat loads. The Residual Forecaster uses these residuals as its target context, with ambient temperature and the interpolated base prediction as covariates. Future inputs contain the corresponding temperature and base predictions.

MRRC's configuration, residual-context selection and evaluation are described in Section~\ref{subsec:mrrc_configuration_evaluation}.

\section{Case Studies and Experimental Setup}\label{sec:data}

This section introduces the Munich and Flensburg DHNs and their heat-load and weather data. Munich is a small, expanding mixed-use network, whereas Flensburg is substantially larger and comparatively stable. Munich's operational data are proprietary; the openly available Flensburg data support reproducible external validation. The section then presents the experimental design, supplementary analyses and MRRC configuration and evaluation.

\subsection{District Heating Networks}\label{subsec:dhn_description}

The data used for model development and primary evaluation originate from a DHN serving a mixed-use urban district in Munich, Germany. At the end of the available measurement period, the network served 11 individual buildings, including large consumers such as hotels and office buildings. The building stock also includes retail, restaurants, cultural and educational facilities, and industrial uses.

Heat is supplied by a centralized energy hub, where the network heat load is measured. The network's supply line is approximately 400~m long. The design supply temperature is 80~$^\circ$C, while the measured supply temperature averaged 84.44~$^\circ$C with a standard deviation of 3.95~$^\circ$C in 2024. The generation portfolio includes two combined heat and power units with a thermal capacity of 1111~kW each, a 1870~kW gas boiler, a 1284~kW high-temperature heat pump, and a 250~kW power-to-heat unit, supported by two 12~$\mathrm{m}^3$ thermal buffer tanks. The combined nominal heat-generation capacity is 5.626 MW. The system is operated using a rule-based control strategy that prioritizes combined heat and power generation, while the heat pump, gas boiler, and power-to-heat unit provide supplementary heat depending on operating conditions.

The available Munich heat-load record spans 1 January 2020 to 31 January 2025 at a native temporal resolution of 15 minutes. Measured weather covariates are obtained from German Weather Service (DWD) stations and aligned with the 15-minute heat load data. Temperature, relative humidity, wind speed, and precipitation are taken from station 03379 M\"unchen-Stadt.\cite{deutscherwetterdienstdwd10minuteStationObservation2026} As this station has no solar irradiation sensor, irradiation data are obtained from station 05404 Weihenstephan-D\"urnast. In 2024, these data were 99.47\% complete; remaining gaps were filled using surrounding DWD stations. In the primary benchmark, these weather variables are treated as known future inputs.

The measured heat-load signal contains occasional periods flagged by the screening procedure described in Section~\ref{subsec:anomaly_screening}. After excluding flagged days, the maximum 15-minute load over the entire available data set was 4.929 MW, and the maximum hourly mean was 3.496 MW.

The Munich network combines consumers with diverse heat-demand characteristics, while its small number of connected buildings limits the smoothing of individual demand fluctuations through aggregation. Together, these features make it a challenging case for district heat-load forecasting. Ongoing expansion adds a further source of change and is accompanied by a substantial increase in annual heat demand, as shown in Supplementary Figure~S1. Table~\ref{tab:munich_distribution_changes} summarizes interannual differences and changes detected within individual years: the Hellinger distance compares each year's heat-load distribution with the preceding year, while the annual number of Kolmogorov--Smirnov Windowing (KSWIN) drift detections indicates how frequently distributional changes were signaled within that year.

\begin{table}[!htbp]
\caption{Distributional-change indicators for the Munich network based on 15-minute heat-load observations. Hellinger distances compare each year’s heat-load distribution with the preceding year; KSWIN drift-detection counts refer to changes detected within the indicated year.}
\label{tab:munich_distribution_changes}
\begin{tabular*}{\tblwidth}{@{\extracolsep{\fill}}lrr@{}}
\toprule
Year & Hellinger distance $H$ & KSWIN drift detections \\
\midrule
2021 & 0.3921 & 65 \\
2022 & 0.1601 & 26 \\
2023 & 0.1827 & 27 \\
2024 & 0.1583 & 31 \\
\bottomrule
\end{tabular*}
\end{table}

\begin{figure}[!htbp]
  \centering
  \includegraphics[width=\linewidth]{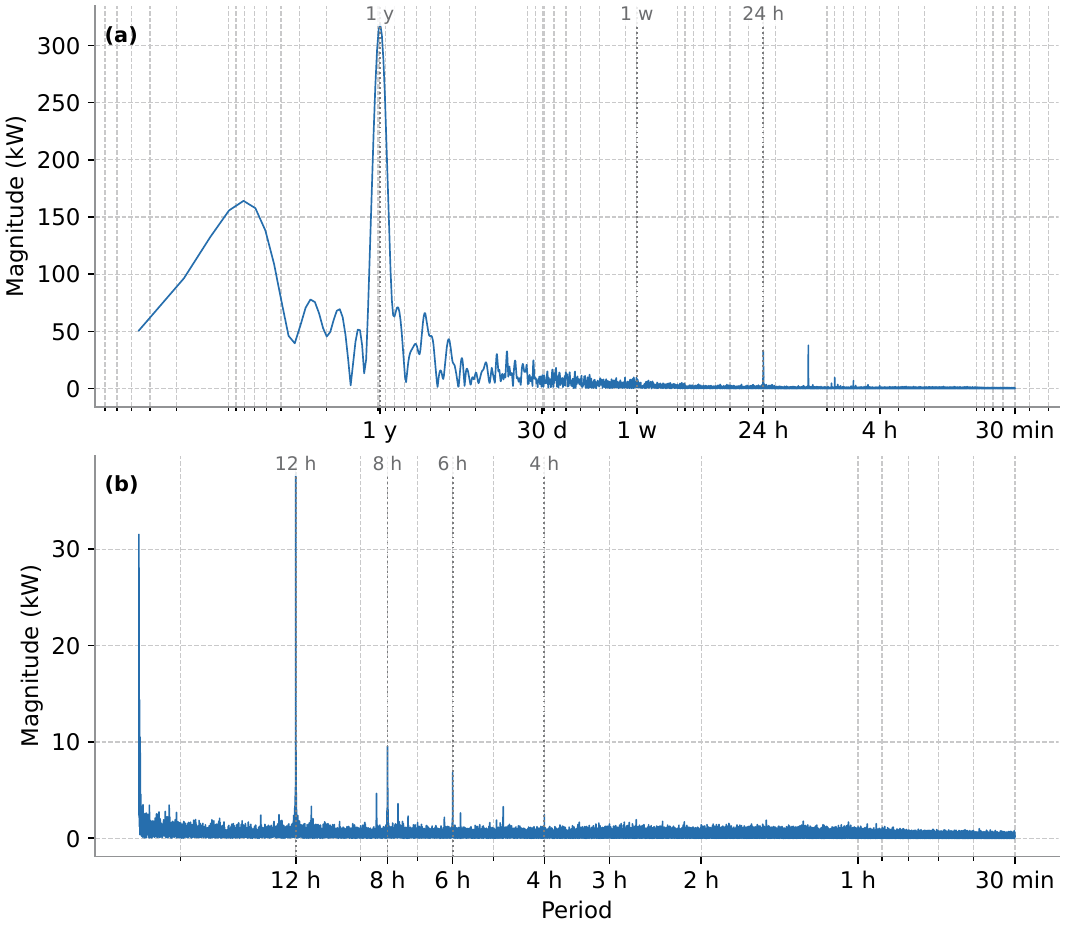}
  \caption{Fourier magnitude spectra of the full available 15-minute heat load time series: (a) full period range and (b) periods up to one day. The x-axis is shown as period rather than frequency; marked periods indicate seasonal, weekly, daily, and sub-daily components.}
  \label{fig:heat_load_fft_combined}
\end{figure}

To characterize the dominant time scales in the measured heat demand, a Fourier spectrum of the full available 15-minute heat load series is shown in Figure~\ref{fig:heat_load_fft_combined}. The very-long-period components above one year should not be interpreted as stationary cycles; they reflect the expansion of the Munich DHN and the associated increase in annual heat demand. Further details of this annual development are provided in Supplementary Figure~S1. The spectrum also shows a dominant seasonal component close to one year, mainly attributable to space-heating demand, and a clear diurnal component around 24~h. Additional peaks at harmonics of the daily cycle, including 12~h, 8~h, 6~h, and 4~h, indicate that the diurnal load pattern contains substantial harmonic content rather than being well described by a single sinusoidal component. The pronounced 12~h component is consistent with two dominant demand periods per day, for example, morning and evening domestic-hot-water peaks. In contrast, no pronounced weekly peak is visible, suggesting that the aggregated DHN signal contains no strong strictly periodic seven-day component.

The data used in the transfer validation originate from the DHN of Flensburg, Germany. The network has developed historically from 1969 into a city-scale district heating system with more than 700 km of district heating pipelines and a connection rate of over 90\% among private households. This is the same network analyzed by Obermeier et al. \cite{obermeier2026fets}, discussed in Section~\ref{subsec:tabpfn_ts_energy_benchmarks}.

Heat is supplied from a central heating plant connected to the DHN. As of 2024, the heat supply infrastructure comprises six generation units and two large thermal storage tanks.\cite{flensburgEinzigartigesFernwaermenetzAnschlussquote} The measured heat load signal reflects the aggregated demand supplied by this city-scale network. For external validation, we use the 2024 hourly heat-load observations from the publicly available 2020--2024 Flensburg district-heating data set.\cite{freissmannNetworkDataDistrict2025} Between 2023, the year supplying the training data for the trained baselines, and 2024, the deployment and evaluation year, Flensburg exhibited a smaller change in its annual heat-load distribution than Munich. Hellinger distances calculated between the annual distributions of hourly average heat load were 0.0780 for Flensburg and 0.1572 for Munich. Weather data for Flensburg are provided by DWD, using hourly air-temperature observations from station 01666 Gl\"ucksburg-Meierwik.

\subsection{Configuration Experiments and Benchmark Evaluation}\label{subsec:experimental_design}

The experimental design comprises configuration diagnostics on selected representative weeks, full-year benchmarks and external validation, and a multi-year evaluation of MRRC. The configuration diagnostics examine context length, temporal resolution, forecast horizon and weather-covariate choice for both foundation models, with additional model-specific diagnostics where appropriate. Both models are then evaluated over 2024 against trained AutoGluon baselines and on the Flensburg data set to assess cross-network transfer. Additional analyses assess sensitivity to anomaly exclusion, baseline retraining following detected concept drift and weather-forecast sensitivity, as described in Sections~\ref{subsec:anomaly_exclusion_setup}--\ref{subsec:weather_sensitivity_setup}. The MRRC configuration and evaluation setup are presented in Section~\ref{subsec:mrrc_configuration_evaluation}.

Although the native temporal resolution of the heat load data is 15 minutes, large thermal buffer tanks partially smooth the network demand. Consequently, this high temporal resolution is not strictly required for heat-generation scheduling. Selected experiments therefore also use hourly aggregated data to assess forecasting performance at a coarser, operationally relevant resolution.

Candidate weeks are restricted to complete Monday--Sunday weeks without observations flagged by the screening procedure in Section~\ref{subsec:anomaly_screening}. From the remaining weeks, three operating regimes are selected based on ambient temperature: the coldest week represents winter operation, the hottest week represents summer operation, and the week with the largest within-week temperature variability represents transitional operation. The selected winter, transitional, and summer weeks in 2024 are January 15--21, April 22--28, and August 12--18, respectively. Their detailed ambient-temperature statistics are provided in Supplementary Table~S1.

These representative weeks are used to study seasonal effects and to compare forecast horizons, context lengths, and temporal resolutions under characteristic operating conditions.

For the principal 2024 benchmarks, the AutoGluon predictors are trained on 2023 target values and evaluated on rolling 24-hour forecasts in 2024. For a transfer validation on a new data set, the AutoGluon models are trained from scratch on the corresponding 2023 data before being evaluated on 2024 data.

For the full-year day-ahead benchmarks, forecasts are issued every 24 elapsed hours, beginning at 00:00 Europe/Berlin on 1 January. This corresponds to 00:00 local time during standard time and 01:00 during daylight-saving time, with each forecast covering 24 elapsed hours. In the selected-week diagnostics, daily forecasts are instead anchored at local midnight at the beginning of each selected week.

\subsection{Sensitivity to Anomaly Exclusion}\label{subsec:anomaly_exclusion_setup}

To assess whether potentially anomalous targets affect the full-year benchmark conclusions, we repeat the evaluation after anomaly exclusion. We exclude complete 24-hour forecasts issued on local dates containing observations flagged by the screening procedure in Section~\ref{subsec:anomaly_screening}.

\subsection{Baseline Retraining on Detected Concept Drift}\label{subsec:drift_retraining_setup}

To test whether batch retraining triggered by active drift detection improves baseline performance, we use KSWIN, which achieved the highest detection accuracy in the benchmark study of Derzsi et al.\cite{derzsiConceptDriftDetection2026} We replay the Munich drift-detection timestamps supplied by the authors, generated using the KSWIN parameters specified for the medium-severity expansion scenario, referred to as EXP-M in Table B.1 of their study.\cite{derzsiConceptDriftDetection2026} AutoGluon models initially fitted on 2020 are evaluated over 2021--2024 either as static models without retraining or with batch retraining from scratch after each KSWIN drift detection on the preceding year of observations, ending strictly before the detection time. At each forecast issuance, the latest eligible model is used; fitting is assumed complete before the next scheduled forecast. Each model receives observed history from its training-window start to the forecast issue time.

\subsection{Weather-Forecast Sensitivity}\label{subsec:weather_sensitivity_setup}

To assess sensitivity to weather-forecast uncertainty, retrospective hourly temperature predictions for Munich were obtained from the Open-Meteo Single Runs API using ECMWF IFS HRES.\cite{zippenfenig_2023_open_meteo} Each 24-hour heat-load forecast uses one coherent temperature trajectory from the preceding 12:00 UTC initialization. The archive contains IFS Cycle 49R1 hindcasts, which we use to assess sensitivity to coherent weather forecasts.

Forecast temperatures are available from 15 March 2024. Requiring a complete 12-week context and 24-hour horizon yields 208 matched daily forecast starts from 7 June to 31 December 2024. The realized-weather reference uses measured temperatures in both the context and future horizon. The forecast-weather variants use the same heat-load histories, forecast starts and targets, but incorporate measured and forecast temperatures as described below.

Measured and forecast temperatures are supplied as separate historical covariates; each historical forecast value comes from the trajectory selected for its corresponding daily forecast start. Chronos-2 receives only forecast temperature as a future covariate. Because the TabPFN-TS implementation requires future values for every covariate, both future columns are filled with the same forecast temperature. These configurations achieved the lowest point-estimate errors among the tested forecast-weather configurations, as shown in Supplementary Table~S2.

\subsection{MRRC Configuration and Evaluation}\label{subsec:mrrc_configuration_evaluation}

We evaluate the MRRC architecture described in Section~\ref{subsec:mrrc_forecaster_method} on the Munich data over 2021--2024 using both TabPFN-TS and Chronos-2 as backbones. The Base Forecaster operates at hourly resolution, and its predictions are linearly interpolated to the 15-minute grid used for residual construction and evaluation. Its forecast horizon is 24 hours, with a new forecast issued every 12 hours. Thus, at every point in time, a forecast for at least the next 12 hours is available. The Residual Forecaster predicts the next two hours at 15-minute resolution and issues a new forecast every hour, ensuring that a high-resolution forecast for at least the next hour is available.

For the Base Forecaster, the context window is chosen based on the 24-hour forecast results described in Section~\ref{subsec:results_context_resolution_horizon}. The base and direct high-resolution forecasts use a common 12-week context to avoid varying context length alongside resolution and update frequency.

Because residual prediction differs from direct heat-load forecasting, the context-window results from the main configuration experiments are not directly transferable to the Residual Forecaster. Residual contexts of 2, 4, 8, 24, 48, 168 and 336 hours are therefore compared separately for each backbone on the held-out December 2020 selection period, preceding the 2021--2024 evaluation. Selection minimizes the combined MRRC forecast's CVRMSE. A seven-day residual context achieved the lowest CVRMSE among the tested context lengths for both TSFM backbones and was selected for evaluation.

We compare the MRRC Forecaster with two baseline setups. The first is the Base Forecaster linearly interpolated to 15-minute resolution without the residual correction. The second baseline is a direct high-frequency, high-resolution predictor: it uses 15-minute resolution, a 13-hour prediction horizon, and hourly re-issuing. This setup ensures that at least 12 hours of forecast are available at every time step while matching the short-term resolution of the MRRC Forecaster. In the following, this setup is referred to as the High Frequency \& High Resolution (HFHR) setup.

Two sets of forecast-quality metrics are reported to reflect the different operational requirements. The short-term metrics assess the latest available 15-minute operational trajectory. For the hourly updated high-resolution and MRRC setups, this corresponds to the first hour after each forecast update and therefore quantifies how well the most recent forecast tracks the realized high-resolution load profile. In contrast, the long-term metrics assess the forecast information available for planning decisions. This long-term evaluation uses the rolling 12-hour energy metric described in Section~\ref{sec:evaluation}. For the 12-hour energy evaluation, the residual correction is applied during the first two hours; the remaining horizon uses the base forecast.

\section{Results}\label{sec:results}

This section presents configuration and seasonal analyses, followed by full-year benchmarks, external validation and probabilistic forecast quality. It then reports sensitivity to anomaly exclusion, KSWIN-triggered baseline retraining and weather-forecast sensitivity, before comparing MRRC with Base and HFHR forecasts across 2021--2024.

\subsection{Configuration Experiments}\label{subsec:results_context_resolution_horizon}

\begin{figure}[!htbp]
  \centering
  \includegraphics[width=\linewidth]{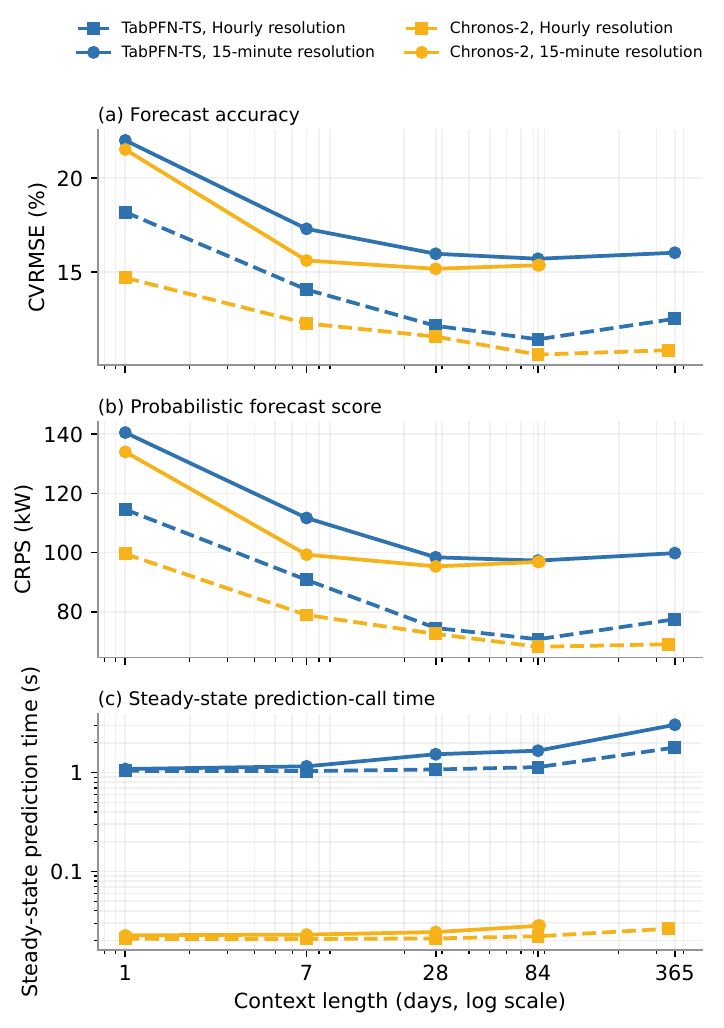}
  \caption{Overall CVRMSE, CRPS, and mean steady-state prediction-call time for 24-hour heat load forecasts as a function of context length for the selected summer, winter, and transitional operating weeks.}
  \label{fig:context_cvrmse_runtime_1day_resolution}
\end{figure}

First, the required length of historical context data was investigated for the 24-hour forecast task. Context windows of 1 day, 1 week, 4 weeks, 12 weeks, and 1 year were evaluated for both 15-minute and hourly forecast resolutions. At hourly resolution, both models achieved their lowest CVRMSE with a 12-week context. At 15-minute resolution, the best context differed: 12 weeks for TabPFN-TS and 4 weeks for Chronos-2, as shown in Figure~\ref{fig:context_cvrmse_runtime_1day_resolution}. The \(R^2\) and MAE results follow the same qualitative trend. For TabPFN-TS, the 12-week context achieved a CVRMSE of 11.42\%, an \(R^2\) of 0.970, and an MAE of 102.6~kW at hourly resolution; at 15-minute resolution, it achieved a CVRMSE of 15.71\%, an \(R^2\) of 0.945, and an MAE of 138.7~kW. Extending the recent context to one year did not further improve the TabPFN-TS point forecast at either resolution. In this experiment Chronos-2 slightly outperforms TabPFN-TS on every metric for all resolutions and context window lengths.

The probabilistic score follows the same main pattern as the deterministic metrics. For hourly forecasts, both models achieved their lowest CRPS with a 12-week context. At 15-minute resolution, the optimum was 12 weeks for TabPFN-TS and four weeks for Chronos-2. For TabPFN-TS, the 12-week context achieved a CRPS of 70.7~kW at hourly resolution and 97.3~kW at 15-minute resolution. Thus, the CRPS results support the choice of a moderate recent context window and do not indicate a probabilistic benefit from using the full one-year context either.

The corresponding steady-state prediction-call times are shown in the lower panel of Figure~\ref{fig:context_cvrmse_runtime_1day_resolution}. For the 12-week context, the mean time per forecast was 1.136~s and 1.665~s for TabPFN-TS at hourly and 15-minute resolution, respectively, compared to 0.022~s and 0.028~s for Chronos-2. Table~\ref{tab:tsfm_controlled_timing} reports model loading and estimated first-call overhead alongside steady-state inference, showing shorter model-loading time for TabPFN-TS but lower first-call overhead and inference time for Chronos-2.

\begin{table}[!htbp]
\centering
\caption{Controlled TSFM timing for hourly 24-hour forecasts with a 12-week context. Values are mean $\pm$ standard deviation across five fresh processes.}
\label{tab:tsfm_controlled_timing}
\begin{tabular*}{\textwidth}{@{\extracolsep{\fill}}lccc@{}}
\toprule
Model & \centeredtablecell{Model loading\\(s)} & \centeredtablecell{Estimated first-call\\overhead (s)} & \centeredtablecell{Steady-state inference\\(s/forecast)} \\
\midrule
TabPFN-TS & $16.49\pm3.45$ & $7.47\pm1.23$ & $0.947\pm0.021$ \\
Chronos-2 & $29.78\pm6.14$ & $0.260\pm0.019$ & $0.01809\pm0.00018$ \\
\bottomrule
\end{tabular*}
\end{table}

\begin{figure}[!htbp]
  \centering
  \includegraphics[width=\linewidth]{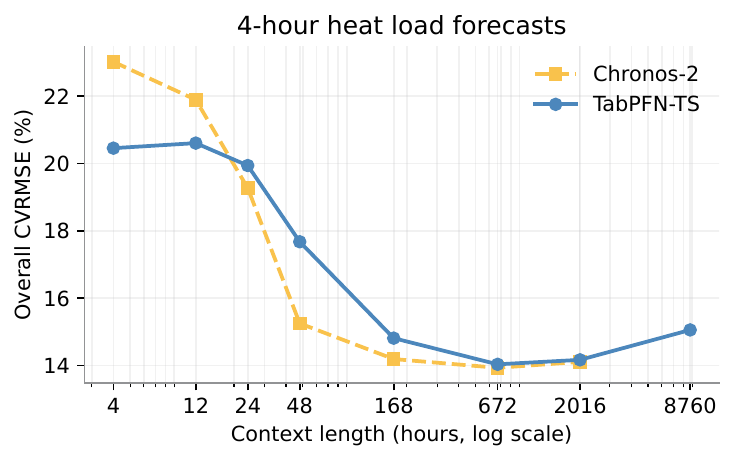}
  \caption{Overall CVRMSE for 4-hour heat load forecasts at 15-minute resolution as a function of context length.}
  \label{fig:cvrmse_context_4h_quarter}
\end{figure}

The influence of the forecast horizon was then evaluated separately. First, a shorter 4-hour horizon was evaluated at 15-minute resolution, with forecasts issued every 4 hours. Figure~\ref{fig:cvrmse_context_4h_quarter} shows how the context-window length affects this short-horizon task. When only very short context windows are available, namely 4 hours or 12 hours, TabPFN-TS outperforms Chronos-2. For medium context windows of 1 day, 2 days, and 1 week, Chronos-2 achieves lower CVRMSE values. For longer context windows of 4 weeks and 12 weeks, both models perform very similarly.

The shorter horizon also changes the context-length optimum for TabPFN-TS. Whereas the 24-hour forecast at 15-minute resolution performed best with a 12-week context window, the 4-hour forecast performed best with a 4-week context window. Compared with the best TabPFN-TS setup for 24-hour forecasts at 15-minute resolution, the best 4-hour setup reduced CVRMSE from 15.71\% to 14.04\%, corresponding to a decrease of 1.68 percentage points, or 10.7\% relative to the 24-hour setup. This comparison reflects the combined effect of the shorter forecast horizon and the higher forecast-update frequency.

To separate the effect of the forecast horizon from the effect of the update frequency, an additional TabPFN-TS run issued 24-hour forecasts every 4 hours and was evaluated with the non-overlapping latest-forecast-wins strategy defined in Section~\ref{sec:evaluation}. With a 12-week context window, this setting achieved a CVRMSE of 14.17\% and an \(R^2\) of 0.955. Its CVRMSE is only 0.13 percentage points higher than the best 4-hour TabPFN-TS forecast, which used a 4-week context window. Thus, most of the improvement over the daily 24-hour 15-minute-resolution forecast is explained by the higher forecast-update frequency. 

\begin{figure}[!htbp]
  \centering
  \includegraphics[width=\linewidth]{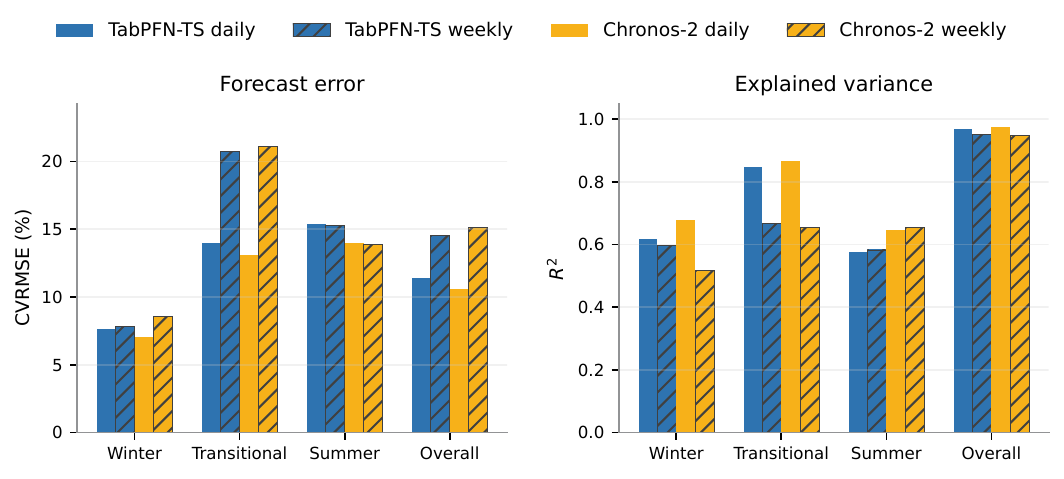}
  \caption{CVRMSE and \(R^2\) for daily 24-hour and weekly 168-hour forecasts at hourly resolution with a 12-week context window for TabPFN-TS and Chronos-2.}
  \label{fig:daily_weekly_horizon_metrics}
\end{figure}

Finally, both models were evaluated for 168-hour forecasts at hourly resolution using a 12-week context. Moving from 24-hour forecasts issued daily to 168-hour forecasts issued weekly reduced accuracy for both models. On the selected operating weeks, TabPFN-TS achieved a slightly lower CVRMSE than Chronos-2, at 14.57\% compared with 15.15\%, with results by operating regime shown in Figure~\ref{fig:daily_weekly_horizon_metrics}. This reversal of the day-ahead ranking motivated an additional full-year evaluation to assess whether the advantage persisted across the year. The full-year results did not confirm the TabPFN-TS advantage: Chronos-2 achieved a lower weekly CVRMSE of 15.64\%, compared with 17.06\% for TabPFN-TS, as reported in Supplementary Table~S3.

\begin{figure}[!htbp]
  \centering
  \includegraphics[width=\textwidth]{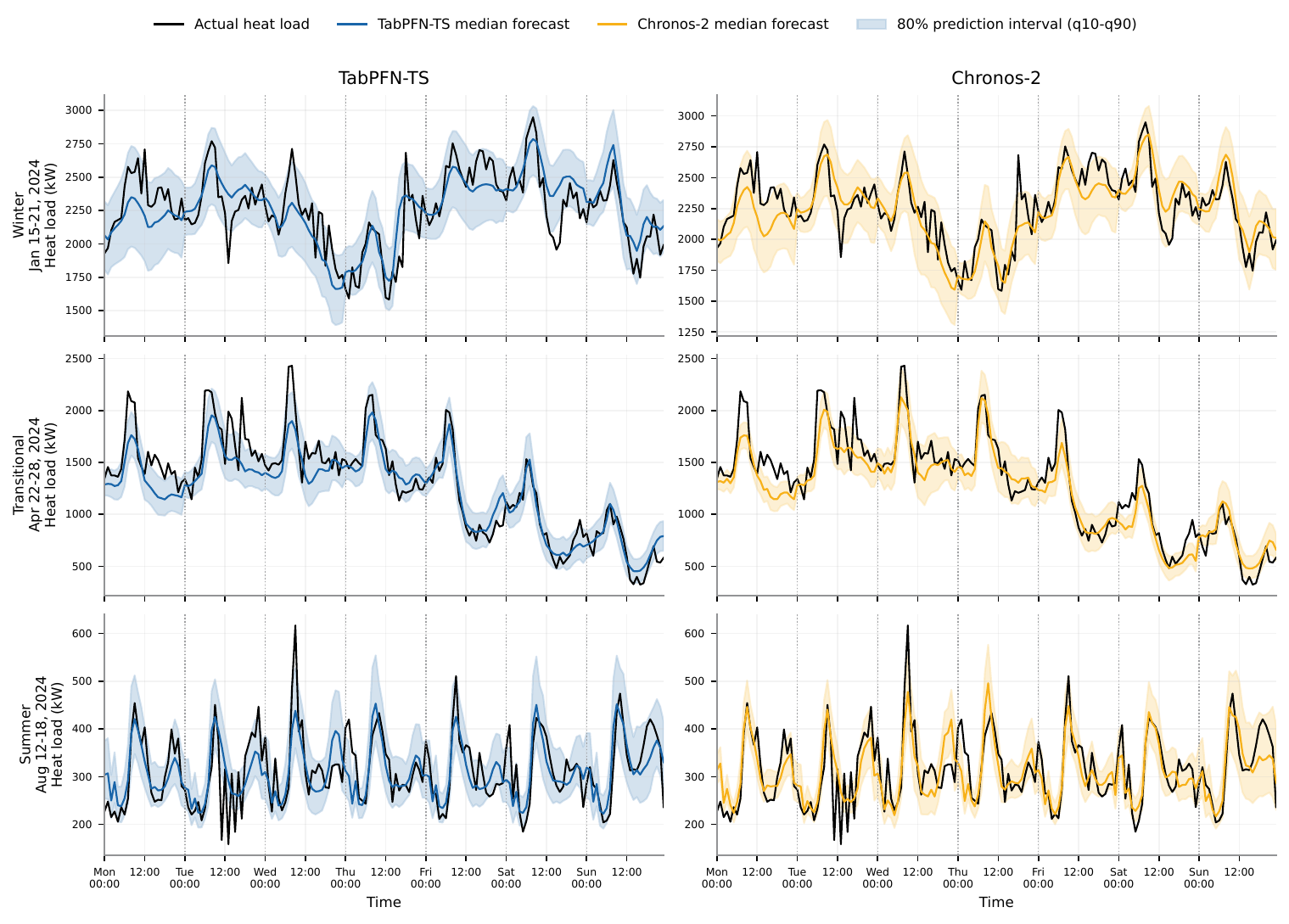}
  \caption{Munich predictions for the selected 24-hour forecast setup at hourly resolution with a 12-week context window.}
  \label{fig:prediction_check_munich_hourly_pred24h_context12w_q10_q90}
\end{figure}

After the context-length, resolution, and horizon diagnostics, the sensitivity to weather-covariate choice was evaluated for the hourly 24-hour forecast setup with a 12-week context window. Table~\ref{tab:weather_feature_selection} reports the aggregate performance over all selected operating weeks.

\begin{table}[!htbp]
\caption{Weather feature selection for 24-hour hourly forecasts with a 12-week context window for TabPFN-TS and Chronos-2.}\label{tab:weather_feature_selection}
\scriptsize
\begin{tabular*}{\tblwidth}{@{\extracolsep{\fill}}LRRRRRR@{}}
\toprule
& \multicolumn{3}{c}{TabPFN-TS} & \multicolumn{3}{c}{Chronos-2} \\
\cmidrule(lr){2-4}\cmidrule(l){5-7}
Weather covariates & CVRMSE (\%) & $R^2$ & MAE (kW) & CVRMSE (\%) & $R^2$ & MAE (kW) \\
\midrule
Amb. temp. & 11.42 & 0.970 & 102.6 & 10.60 & 0.974 & 96.0 \\
Amb. temp., relative humidity & 11.43 & 0.970 & 102.2 & 10.62 & 0.974 & 96.7 \\
Amb. temp., wind speed & 11.09 & 0.972 & 98.8 & 10.56 & 0.975 & 96.0 \\
Amb. temp., precipitation & 11.16 & 0.972 & 100.1 & 10.59 & 0.975 & 96.0 \\
Amb. temp., precip., wind sp. & 11.25 & 0.971 & 99.9 & 10.38 & 0.976 & 94.2 \\
Amb. temp., solar irradiation & 11.33 & 0.971 & 101.5 & 10.47 & 0.975 & 95.3 \\
All weather covariates & 11.09 & 0.972 & 98.8 & 10.32 & 0.976 & 95.6 \\
\bottomrule
\end{tabular*}
\end{table}

Additional weather covariates produced small, model-dependent improvements on the selected weeks. Ambient temperature alone was retained as a parsimonious configuration. It is measured at many sites and is commonly available as a forecast variable, supporting transferability and operational simplicity. The mean prediction-call time increases by 2.4\% for TabPFN-TS and 3.3\% for Chronos-2 when moving from ambient temperature only to all weather covariates.

Motivated by the dominant annual spectral peak in the Munich heat-load data described in Section~\ref{subsec:dhn_description}, we examine whether augmenting recent context with calendar-aligned observations from the previous year improves TabPFN-TS forecast accuracy.

Here, relevant observations denote historical data from a window around the current calendar date, shifted by one year. The recent-plus-relevant setup combines 6 weeks of recent context with a 6-week block from the same calendar period one year earlier, so that the total amount of context data matches the observed optimal 12-week recent-data context. According to the TabPFN-TS documentation, automatic temporal feature engineering must be disabled for this non-contiguous context configuration. Therefore, the recent-plus-relevant experiment used only calendar features and disabled the running-index and automatic seasonal features. This experiment was carried out only for TabPFN-TS, because Chronos-2 expects strictly regular time steps without gaps in the context. 

No performance benefit was obtained from adding a 6-week block from the same calendar period one year earlier. Table~\ref{tab:recent_relevant_context} also shows that the default automatic temporal feature engineering reduced CVRMSE by 2.3\% compared with the recent-only setup.

\begin{table}[!htbp]
\caption{Context-data selection for TabPFN-TS hourly 24-hour forecasts. Context lengths are in weeks; ``auto feat.'' denotes automatic temporal features.}\label{tab:recent_relevant_context}
\footnotesize
\begin{tabular*}{\tblwidth}{@{\extracolsep{\fill}}LRRR@{}}
\toprule
Setup & CVRMSE (\%) & $R^2$ & MAE (kW) \\
\midrule
Recent 12 (auto feat.) & 11.42 & 0.970 & 102.6 \\
Recent 12 & 11.69 & 0.969 & 106.9 \\
Recent 6 + Relevant 6 & 12.69 & 0.963 & 113.0 \\
\bottomrule
\end{tabular*}
\end{table}

Taken together, the context-length, temporal-resolution, forecast-horizon, weather-covariate, and context-data diagnostics motivate using 24-hour-ahead forecasts at hourly resolution with a 12-week rolling context window and ambient temperature as the only weather covariate in the subsequent full-year and transfer analyses. Figure~\ref{fig:prediction_check_munich_hourly_pred24h_context12w_q10_q90} shows the resulting TabPFN-TS and Chronos-2 forecasts for the selected winter, transitional, and summer weeks.

\subsection{Seasonal Effects on Forecast Accuracy and Uncertainty}\label{sec:seasonal_effects}

\begin{table}[!htbp]
\caption{CRPS and 80\% prediction-interval widths for daily issued forecasts at hourly resolution with a 12-week context on the selected representative weeks.}\label{tab:prediction_interval_quality}
\footnotesize
\begin{tabular*}{\tblwidth}{@{\extracolsep{\fill}}LLRRR@{}}
\toprule
Period & Model & CRPS (kW) & Width (kW) & Rel. w. (\%) \\
\midrule
Winter & TabPFN-TS & 92.95 & 498.9 & 22.1 \\
 & Chronos-2 & 85.56 & 488.4 & 21.6 \\
\midrule
Transitional & TabPFN-TS & 93.81 & 370.3 & 28.5 \\
 & Chronos-2 & 95.20 & 317.7 & 24.5 \\
\midrule
Summer & TabPFN-TS & 25.42 & 114.8 & 36.7 \\
 & Chronos-2 & 23.78 & 93.9 & 30.0 \\
\midrule
Overall & TabPFN-TS & 70.73 & 328.0 & 25.4 \\
 & Chronos-2 & 68.18 & 300.0 & 23.2 \\
\bottomrule
\end{tabular*}
\end{table}

Figure~\ref{fig:daily_weekly_horizon_metrics} shows that forecast performance depends strongly on the operating period. For daily 24-hour forecasts, the CVRMSE is lowest in winter, followed by the transitional period and summer. The \(R^2\) score follows a different pattern: winter and summer are comparable, while the transitional week achieves the highest \(R^2\). During the transitional week, the loss in accuracy when moving from daily to weekly forecasts is particularly visible.

The seasonal pattern is also reflected in the probabilistic forecasts. Table~\ref{tab:prediction_interval_quality} reports CRPS and the absolute and relative widths of the 80\% prediction interval. TabPFN-TS has higher CRPS than Chronos-2 in winter, in summer, and across the three weeks combined, indicating lower overall probabilistic forecast quality under this score. During the transitional week, however, TabPFN-TS achieves lower CRPS than Chronos-2, at 93.81 versus 95.20~kW. TabPFN-TS produces wider prediction intervals in all three weeks. For both models, the relative interval width is lowest in winter and highest in summer, with the transitional period in between. During the selected transitional week, both models show a systematic tendency to underestimate heat demand, particularly Chronos-2. The seasonal PIT histograms in Supplementary Figure~S2 provide further detail.

\subsection{Full-Year Benchmark and Transfer Validation}\label{subsec:results_full_year_transfer}

\begin{table}[!htbp]
\caption{Full-year benchmark and transfer-validation results for hourly 24-hour forecasts in 2024. Brackets denote 95\% paired seven-day block-bootstrap confidence intervals.}\label{tab:full_year_transfer_benchmark}
\scriptsize
\newcommand{\benchmarkcell}[1]{\begin{tabular}[c]{@{}c@{}}#1\end{tabular}}
\begin{tabular*}{\tblwidth}{@{\extracolsep{\fill}}LLRRRRRR@{}}
\toprule
& & \multicolumn{3}{c}{Aggregate metrics} & \multicolumn{3}{c}{Mean daily rank} \\
\cmidrule(lr){3-5}\cmidrule(l){6-8}
Model & Rank shift & CVRMSE (\%) & $R^2$ & MAE (kW) & CVRMSE & $R^2$ & MAE \\
\midrule
\multicolumn{8}{@{}l}{\textit{Munich}} \\
Chronos-2 &  & \benchmarkcell{12.48\\{[10.88, 14.37]}} & \benchmarkcell{0.962\\{[0.946, 0.972]}} & \benchmarkcell{85.4\\{[75.9, 95.6]}} & 3.04 & 3.01 & 3.11 \\
TabPFN-TS &  & \benchmarkcell{13.07\\{[11.49, 14.94]}} & \benchmarkcell{0.959\\{[0.941, 0.969]}} & \benchmarkcell{90.3\\{[79.9, 101.1]}} & 3.89 & 3.86 & 3.83 \\
TFT &  & \benchmarkcell{14.61\\{[12.92, 16.61]}} & \benchmarkcell{0.948\\{[0.928, 0.961]}} & \benchmarkcell{102.1\\{[90.7, 114.3]}} & 5.88 & 5.85 & 5.93 \\
Direct LGBM &  & \benchmarkcell{15.87\\{[14.25, 17.72]}} & \benchmarkcell{0.939\\{[0.917, 0.953]}} & \benchmarkcell{113.7\\{[99.1, 129.6]}} & 6.49 & 6.42 & 6.49 \\
DeepAR &  & \benchmarkcell{16.54\\{[14.89, 18.35]}} & \benchmarkcell{0.934\\{[0.910, 0.949]}} & \benchmarkcell{117.2\\{[101.1, 134.1]}} & 6.69 & 6.64 & 6.75 \\
AutoGluon Ensemble &  & \benchmarkcell{16.54\\{[14.89, 18.35]}} & \benchmarkcell{0.934\\{[0.910, 0.949]}} & \benchmarkcell{117.2\\{[101.1, 134.1]}} & 6.69 & 6.64 & 6.75 \\
Direct XGBoost &  & \benchmarkcell{17.34\\{[15.59, 19.30]}} & \benchmarkcell{0.927\\{[0.904, 0.942]}} & \benchmarkcell{126.7\\{[110.6, 144.9]}} & 8.44 & 8.39 & 8.61 \\
AutoETS &  & \benchmarkcell{18.43\\{[16.31, 20.90]}} & \benchmarkcell{0.918\\{[0.884, 0.938]}} & \benchmarkcell{132.8\\{[117.1, 148.9]}} & 7.75 & 7.73 & 7.84 \\
ETS &  & \benchmarkcell{18.43\\{[16.31, 20.90]}} & \benchmarkcell{0.918\\{[0.884, 0.938]}} & \benchmarkcell{132.8\\{[117.1, 148.9]}} & 7.75 & 7.73 & 7.84 \\
DLinear &  & \benchmarkcell{18.66\\{[16.58, 21.06]}} & \benchmarkcell{0.916\\{[0.882, 0.937]}} & \benchmarkcell{135.1\\{[118.6, 151.9]}} & 8.67 & 8.69 & 8.75 \\
Feed-forward &  & \benchmarkcell{19.40\\{[17.33, 21.86]}} & \benchmarkcell{0.909\\{[0.874, 0.931]}} & \benchmarkcell{141.9\\{[124.5, 159.4]}} & 9.59 & 9.63 & 9.67 \\
AutoARIMA &  & \benchmarkcell{20.48\\{[18.13, 23.24]}} & \benchmarkcell{0.898\\{[0.857, 0.924]}} & \benchmarkcell{148.3\\{[129.8, 167.0]}} & 10.09 & 10.12 & 9.88 \\
Recursive LGBM &  & \benchmarkcell{21.07\\{[18.67, 23.91]}} & \benchmarkcell{0.892\\{[0.848, 0.920]}} & \benchmarkcell{152.0\\{[132.7, 171.7]}} & 10.63 & 10.69 & 10.52 \\
Seasonal naive &  & \benchmarkcell{22.51\\{[19.94, 25.58]}} & \benchmarkcell{0.877\\{[0.826, 0.909]}} & \benchmarkcell{161.9\\{[141.1, 183.1]}} & 11.64 & 11.72 & 11.52 \\
Naive &  & \benchmarkcell{23.66\\{[20.75, 27.22]}} & \benchmarkcell{0.864\\{[0.803, 0.901]}} & \benchmarkcell{172.1\\{[151.8, 193.8]}} & 12.77 & 12.89 & 12.52 \\
\midrule
\multicolumn{8}{@{}l}{\textit{Flensburg}} \\
Chronos-2 &  & \benchmarkcell{9.82\\{[8.77, 11.14]}} & \benchmarkcell{0.970\\{[0.957, 0.977]}} & \benchmarkcell{7913.6\\{[6828.9, 9058.9]}} & 3.98 & 3.98 & 4.14 \\
TabPFN-TS &  & \benchmarkcell{10.37\\{[9.30, 11.71]}} & \benchmarkcell{0.966\\{[0.952, 0.975]}} & \benchmarkcell{8437.3\\{[7265.3, 9664.9]}} & 4.77 & 4.77 & 4.60 \\
TFT &  & \benchmarkcell{10.69\\{[9.36, 12.27]}} & \benchmarkcell{0.964\\{[0.948, 0.974]}} & \benchmarkcell{8669.2\\{[7347.2, 10123.4]}} & 5.07 & 5.07 & 5.06 \\
DeepAR & \(\uparrow 1\) & \benchmarkcell{10.72\\{[9.68, 11.91]}} & \benchmarkcell{0.964\\{[0.950, 0.972]}} & \benchmarkcell{8677.9\\{[7363.6, 10059.6]}} & 5.01 & 5.01 & 5.12 \\
AutoGluon Ensemble & \(\uparrow 1\) & \benchmarkcell{10.72\\{[9.68, 11.91]}} & \benchmarkcell{0.964\\{[0.950, 0.972]}} & \benchmarkcell{8677.9\\{[7363.6, 10059.6]}} & 5.01 & 5.01 & 5.12 \\
Direct LGBM & \(\downarrow 2\) & \benchmarkcell{11.12\\{[10.09, 12.27]}} & \benchmarkcell{0.961\\{[0.947, 0.970]}} & \benchmarkcell{9380.2\\{[8098.7, 10709.1]}} & 6.20 & 6.20 & 6.50 \\
Direct XGBoost &  & \benchmarkcell{11.43\\{[10.39, 12.61]}} & \benchmarkcell{0.959\\{[0.944, 0.968]}} & \benchmarkcell{9716.2\\{[8396.1, 11066.6]}} & 6.89 & 6.89 & 7.07 \\
Feed-forward & \(\uparrow 3\) & \benchmarkcell{13.05\\{[11.97, 14.35]}} & \benchmarkcell{0.946\\{[0.926, 0.959]}} & \benchmarkcell{10665.8\\{[9103.6, 12276.0]}} & 7.44 & 7.44 & 7.53 \\
DLinear & \(\uparrow 1\) & \benchmarkcell{13.62\\{[12.59, 14.84]}} & \benchmarkcell{0.941\\{[0.921, 0.954]}} & \benchmarkcell{11200.1\\{[9568.3, 12896.1]}} & 8.64 & 8.64 & 8.58 \\
Recursive LGBM & \(\uparrow 3\) & \benchmarkcell{13.75\\{[12.68, 14.92]}} & \benchmarkcell{0.940\\{[0.921, 0.953]}} & \benchmarkcell{11456.8\\{[9835.7, 13095.7]}} & 9.23 & 9.23 & 9.22 \\
Seasonal naive & \(\uparrow 3\) & \benchmarkcell{15.53\\{[14.19, 17.12]}} & \benchmarkcell{0.924\\{[0.896, 0.942]}} & \benchmarkcell{12775.2\\{[10905.9, 14708.0]}} & 10.68 & 10.68 & 10.40 \\
AutoARIMA &  & \benchmarkcell{15.80\\{[14.36, 17.54]}} & \benchmarkcell{0.921\\{[0.890, 0.941]}} & \benchmarkcell{12820.7\\{[10811.6, 14875.6]}} & 9.66 & 9.66 & 9.55 \\
AutoETS & \(\downarrow 5\) & \benchmarkcell{21.21\\{[20.12, 22.30]}} & \benchmarkcell{0.858\\{[0.813, 0.886]}} & \benchmarkcell{18255.5\\{[15406.1, 21185.2]}} & 12.43 & 12.43 & 12.38 \\
ETS & \(\downarrow 5\) & \benchmarkcell{21.21\\{[20.12, 22.30]}} & \benchmarkcell{0.858\\{[0.813, 0.886]}} & \benchmarkcell{18255.5\\{[15406.1, 21185.2]}} & 12.43 & 12.43 & 12.38 \\
Naive &  & \benchmarkcell{22.39\\{[21.19, 23.57]}} & \benchmarkcell{0.842\\{[0.791, 0.874]}} & \benchmarkcell{18766.4\\{[15467.8, 22171.0]}} & 12.56 & 12.56 & 12.34 \\
\bottomrule
\end{tabular*}
\end{table}

\begin{figure}[!htbp]
  \centering
  \begin{minipage}{0.49\textwidth}
    \centering
    \includegraphics[width=\linewidth]{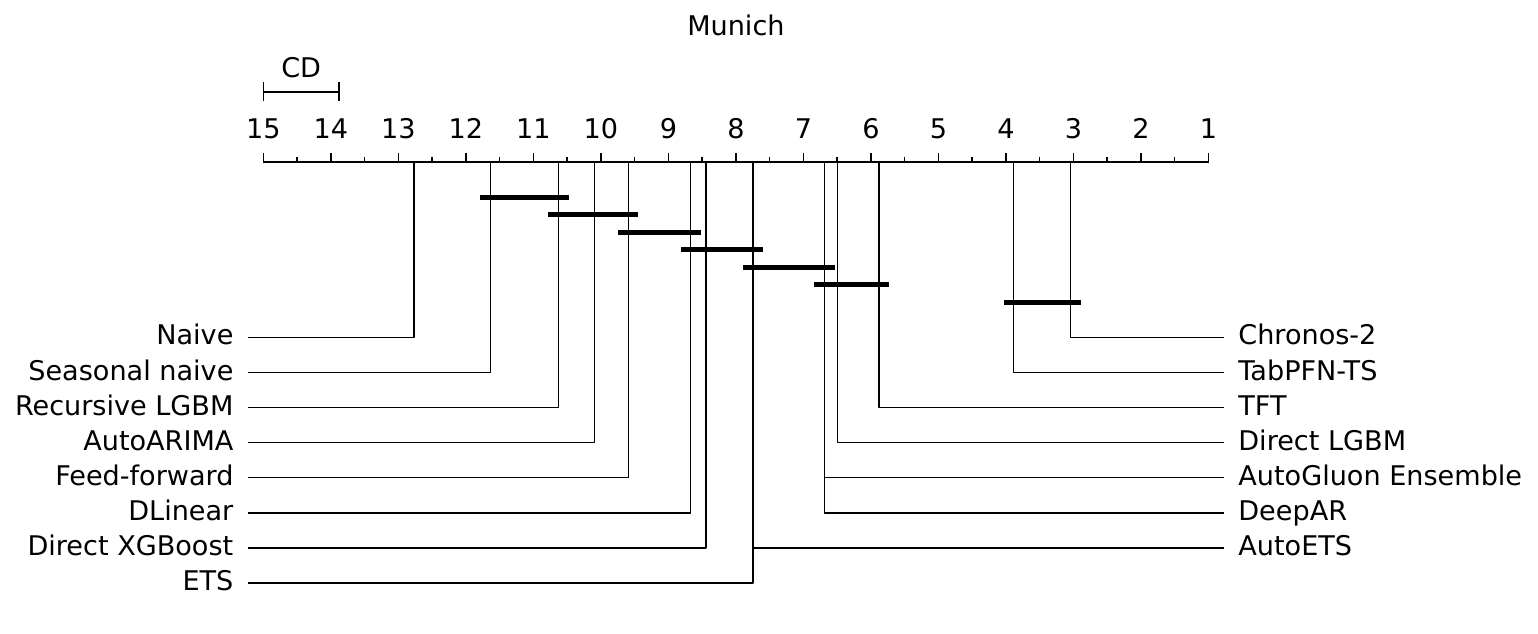}
  \end{minipage}\hfill
  \begin{minipage}{0.49\textwidth}
    \centering
    \includegraphics[width=\linewidth]{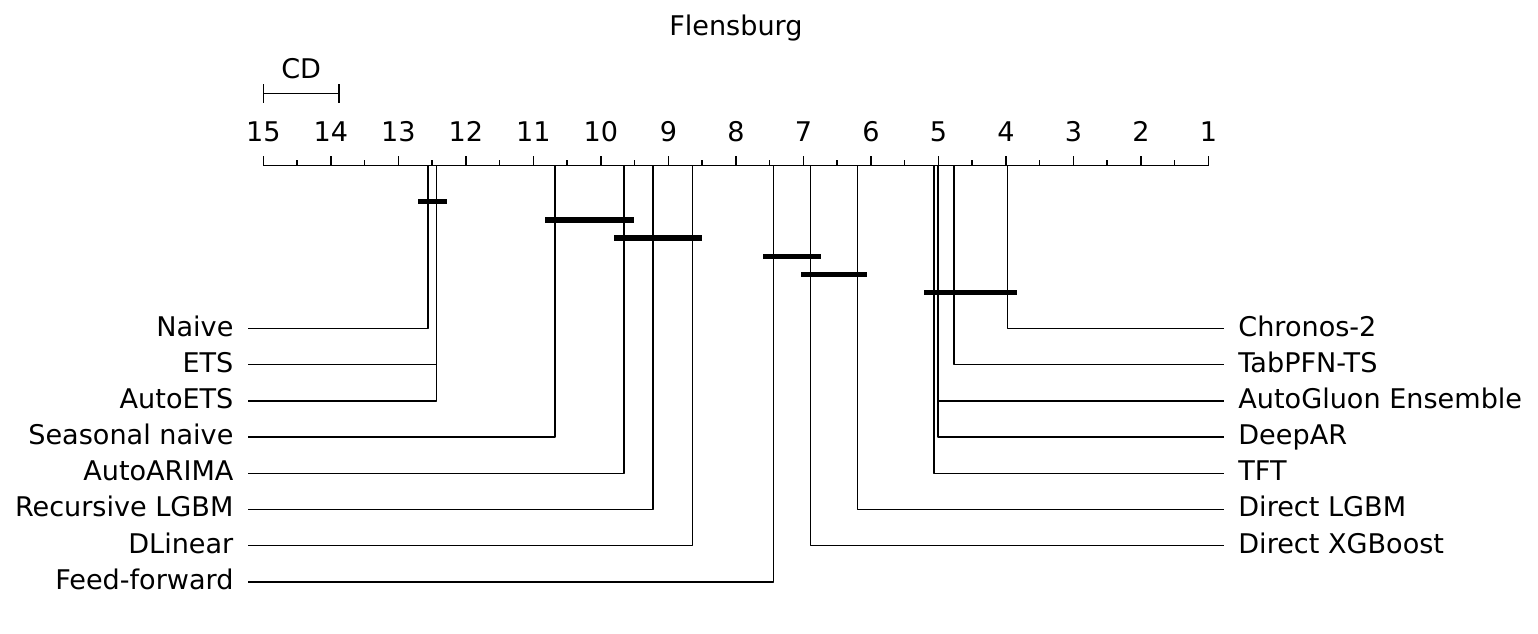}
  \end{minipage}
  \par\medskip
  \includegraphics[width=\textwidth]{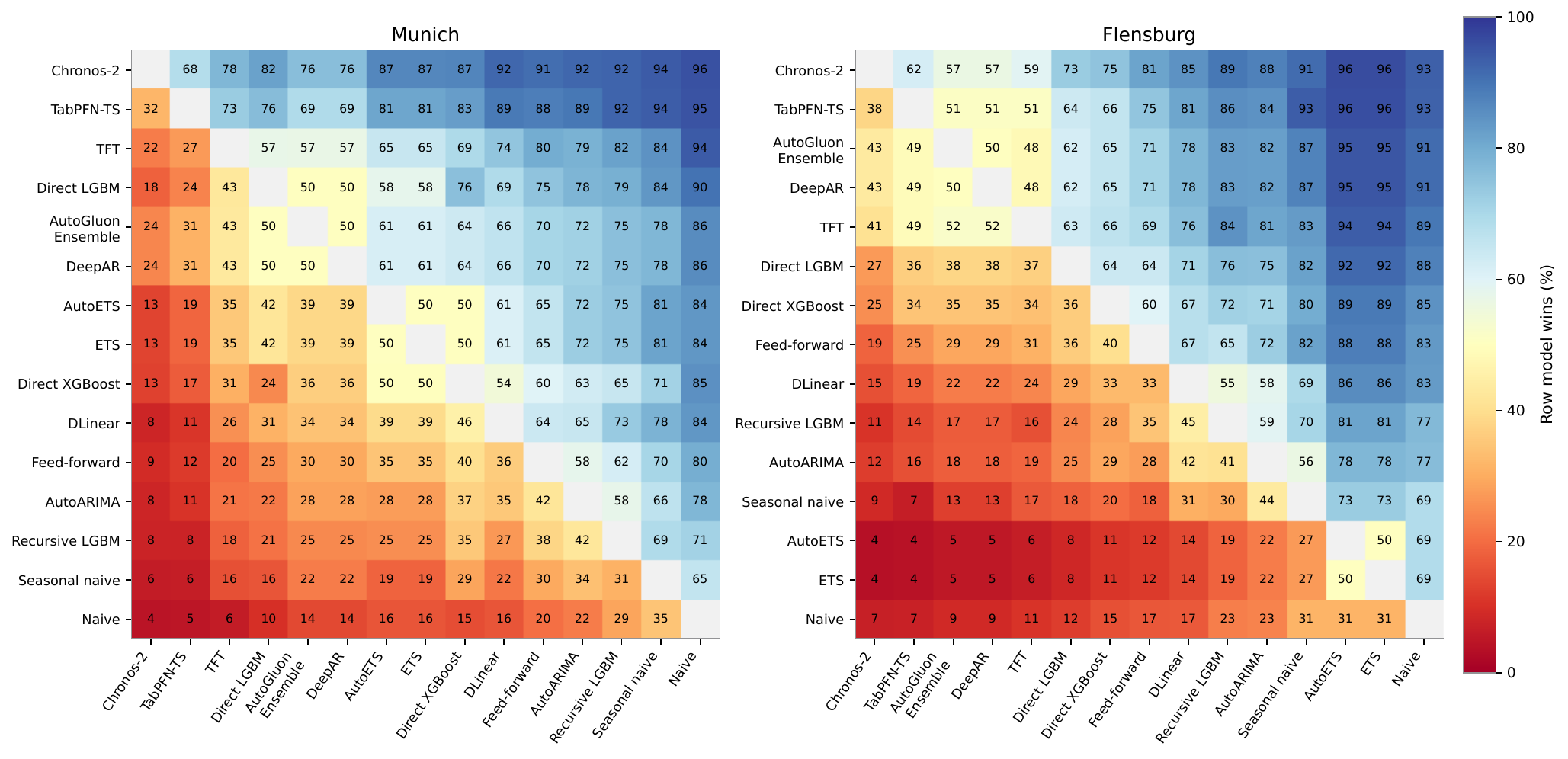}
  \caption{Full-year model comparison based on daily CVRMSE. The upper panels show CD diagrams and the lower panels show pairwise win rates. Each heatmap cell gives the percentage of forecast days on which the row model outperforms the column model; exact ties count as half a win.}
  \label{fig:critical_difference_model_comparison}
\end{figure}

The results of the full-year simulations are reported in Table~\ref{tab:full_year_transfer_benchmark}. The entries are sorted by aggregate CVRMSE within each network. On the Munich data set, both foundation models outperformed the trained AutoGluon baselines across the reported accuracy metrics, with Chronos-2 achieving the lowest aggregate error. To assess whether models perform consistently when transferred to a different data set, the rank-shift column reports the change in CVRMSE position in Flensburg relative to the Munich ordering. The relative error metrics were generally lower for Flensburg than for Munich, especially for the strongest models. Chronos-2, TabPFN-TS, and TFT remained the three strongest models in both networks, whereas the naive baselines stayed at the lower end of the ranking. The largest rank changes occurred for AutoETS and ETS, which lost five CVRMSE positions in Flensburg, and for Feed-forward, Recursive LGBM, and Seasonal naive, which gained three positions each. Chronos-2 again achieved the lowest aggregate error, while TabPFN-TS remained ahead of the trained AutoGluon models.

The critical-difference diagrams are shown in Figure~\ref{fig:critical_difference_model_comparison}. For both data sets, TabPFN-TS lies within the critical difference of Chronos-2. On the Munich data set, the two time-series foundation models form the leading group and outperform all trained AutoGluon baselines in the daily-rank comparison. In Flensburg, the leading group is broader: TFT, DeepAR, and AutoGluon Ensemble also lie within the critical difference of the foundation-model results. The lower panels of Figure~\ref{fig:critical_difference_model_comparison} show the corresponding pairwise daily CVRMSE win rates. Supplementary Table~S8 reports initial AutoGluon fitting times.

\subsection{Probabilistic Forecast Quality}\label{subsec:results_probabilistic_quality}

\begin{figure}[!htbp]
  \centering
  \includegraphics[width=0.95\textwidth]{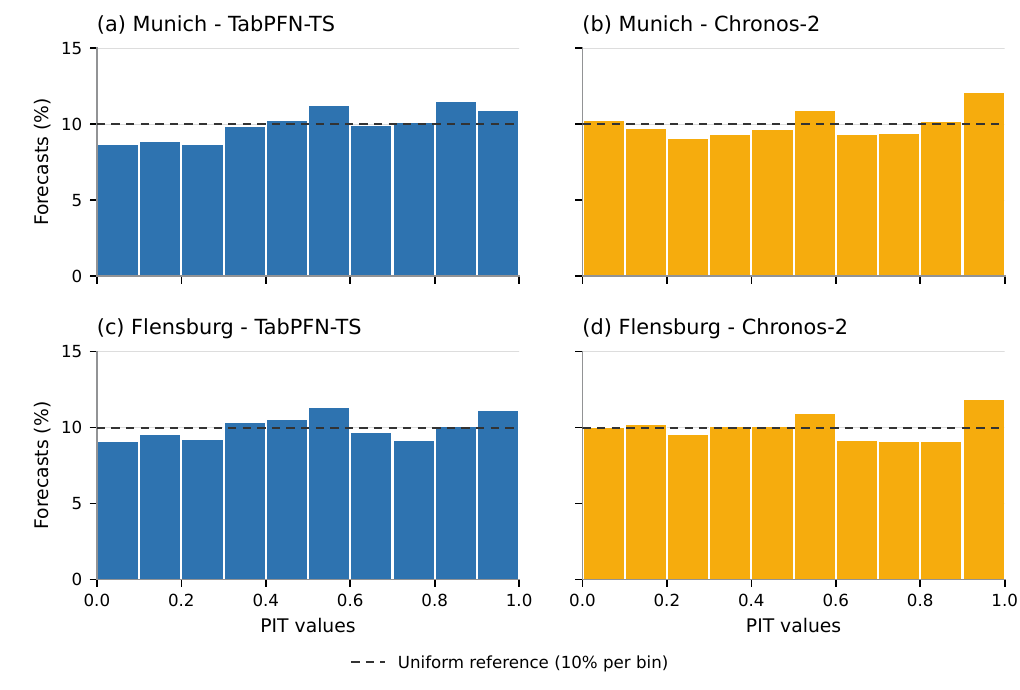}
  \caption{Approximate PIT histograms for the principal 2024 hourly, 24-hour forecasting benchmark in Munich and Flensburg.}
  \label{fig:full_year_probabilistic_scores}
\end{figure}

Chronos-2 achieves lower full-year CRPS than TabPFN-TS in both networks: 60.49 versus 63.66 kW in Munich and 5.62 versus 5.94 MW in Flensburg, indicating better overall probabilistic forecast quality. Figure~\ref{fig:full_year_probabilistic_scores} compares the approximate PIT distributions. Both models show deviations from uniformity. In Munich, TabPFN-TS shows a broader shift towards higher PIT values, suggesting a general tendency to underpredict heat load. Chronos-2 instead exhibits an excess concentrated in the highest PIT bin in both networks, indicating that its upper predictive tail does not adequately cover the observed loads.

\subsection{Sensitivity to Anomaly Exclusion}\label{subsec:results_anomaly_exclusion}

We also investigated the influence of potentially anomalous target observations using the screening criteria described in Section~\ref{subsec:anomaly_screening}. After excluding 24 Munich forecast days containing time steps identified by these criteria, 342 evaluation days remained. The complete aggregate CVRMSE ordering was unchanged, with Chronos-2, TabPFN-TS and TFT retaining the first three positions. Chronos-2 and TabPFN-TS also remained the leading group in the critical-difference comparison. CVRMSE and MAE decreased for 14 of the 15 models; the naive baseline was the exception. For TabPFN-TS and Chronos-2, the standard deviation of daily MAE decreased from 51.8 to 48.8~kW and from 49.4 to 45.0~kW, respectively. Both foundation models achieved lower CRPS after exclusion, with Chronos-2 retaining the lower score. No Flensburg day met the screening criteria. Supplementary Table~S4 reports the detailed deterministic results in Panel A and paired changes after anomaly exclusion for all 15 models in Panel B. Full-year and anomaly-filtered CRPS values are reported in Supplementary Table~S5.

\subsection{Baseline Retraining on Detected Concept Drift}\label{subsec:results_drift_retraining}

Table~\ref{tab:kswin_annual_cvrmse} compares static models and models undergoing KSWIN-triggered batch retraining over 2021--2024, averaging scores equally across the trained AutoGluon setups: TFT, DeepAR, DLinear, Feed-forward, Direct LGBM, Direct XGBoost, Recursive LGBM, and the AutoGluon Ensemble. Retraining improves their average performance in every year, with paired confidence intervals excluding zero. The relative CVRMSE reduction increases continuously from 4.4\% in 2021 to 18.4\% in 2024 as the deployment period moves further from the initial 2020 training period. MAE and $R^2$ show the same direction of improvement and increasing relative gains, as shown in Supplementary Table~S6.

\begin{table}[!htbp]
\caption{Annual mean CVRMSE across eight learned AutoGluon setups in Munich. Differences are KSWIN minus static; brackets denote 95\% percentile confidence intervals.}\label{tab:kswin_annual_cvrmse}
\small
\begin{tabular*}{\tblwidth}{@{\extracolsep{\fill}}ccccc@{}}
\toprule
Year & \centeredtablecell{Static CVRMSE\\(\%)} & \centeredtablecell{KSWIN CVRMSE\\(\%)} & \centeredtablecell{$\Delta$CVRMSE\\(percentage points)} & \centeredtablecell{Relative reduction\\(\%)} \\
\midrule
2021 & 19.96 [18.24, 22.10] & 19.09 [17.27, 21.34] & $-0.88$ [$-1.42$, $-0.28$] & 4.4 [1.4, 7.2] \\
2022 & 19.55 [17.18, 22.68] & 17.62 [14.96, 21.31] & $-1.93$ [$-2.77$, $-1.02$] & 9.9 [4.8, 14.8] \\
2023 & 22.85 [20.40, 25.60] & 19.67 [17.18, 22.73] & $-3.19$ [$-4.08$, $-2.20$] & 13.9 [9.3, 18.1] \\
2024 & 20.69 [19.00, 22.49] & 16.89 [15.09, 18.93] & $-3.80$ [$-4.87$, $-2.65$] & 18.4 [12.7, 23.4] \\
\bottomrule
\end{tabular*}
\end{table}

The benefit is model-dependent. Over the complete four-year period, Direct XGBoost and Direct LGBM show the largest CVRMSE reductions, from 30.03\% to 19.31\% and from 26.21\% to 18.81\%, respectively, whereas DLinear and Feed-forward deteriorate after retraining, as detailed in Supplementary Table~S7. The annual results do not show a consistent relationship between the refitting benefit and either the Hellinger distance or the number of KSWIN drift detections in Table~\ref{tab:munich_distribution_changes}.

The baseline ranking also changes relative to the main 2024 benchmark trained on 2023 data: The static AutoGluon Ensemble ranks ahead of TFT over 2021--2024, while Direct LGBM falls behind several other baselines. Details can be found in Supplementary Table~S7. This illustrates the dependence of trained-model performance on the training and deployment periods. Across the evaluated training setups and evaluation periods, Chronos-2 and TabPFN-TS retain the leading point-estimate performance, with and without KSWIN-triggered baseline retraining. The annual CVRMSE gap between the mean of the eight static trained-baseline setups and each TSFM increases monotonically from 2021 to 2024: from 4.73 to 7.62 percentage points for TabPFN-TS and from 5.42 to 8.21 percentage points for Chronos-2. With KSWIN-triggered retraining, the corresponding gaps remain within 3.31--3.86 and 3.66--4.54 percentage points, respectively, without a monotonic increase.

Retraining the complete AutoGluon model suite took an average of 179.72~s per refit. Supplementary Table~S9 reports the cumulative refitting time divided by prediction-call time for each model.

\subsection{Weather-Forecast Sensitivity}\label{subsec:results_forecast_uncertainty}

Table~\ref{tab:weather_forecast_sensitivity} summarizes the resulting change in point and probabilistic forecast quality. Both models are affected, with Chronos-2 showing slightly greater sensitivity in this experiment: its CVRMSE increases by 10.1\%, compared with 7.4\% for TabPFN-TS. The degradation is visible for both models in point-forecast accuracy and distributional scores. The paired 95\% confidence intervals for changes in CVRMSE, MAE and $R^2$ exclude zero for both models. Details can be found in Supplementary Section~S3. The standard deviation of daily MAE also increased from 42.0 to 46.2~kW for TabPFN-TS and from 40.0 to 46.7~kW for Chronos-2, indicating greater day-to-day variability in forecast error.

\begin{table}[!htbp]
\caption{Forecast quality with realized temperatures and coherent retrospective ECMWF temperature predictions for 208 matched Munich forecast starts from 7 June to 31 December 2024.}\label{tab:weather_forecast_sensitivity}
\footnotesize
\begin{tabular*}{\tblwidth}{@{\extracolsep{\fill}}LLllll@{}}
\toprule
 & & \multicolumn{3}{c}{Point forecast} & Probabilistic forecast \\
\cmidrule(lr){3-5}\cmidrule(l){6-6}
Model & Weather cov. & CVRMSE (\%) & \(R^2\) & MAE (kW) & CRPS (kW) \\
\midrule
TabPFN-TS & Realized & 11.89 & 0.974 & 72.2 & 50.9 \\
 & Predicted & 12.77 (+7.4\%) & 0.971 (-0.4\%) & 77.8 (+7.7\%) & 55.0 (+8.1\%) \\
\midrule
Chronos-2 & Realized & 11.41 & 0.976 & 68.5 & 48.3 \\
 & Predicted & 12.56 (+10.1\%) & 0.971 (-0.5\%) & 76.1 (+11.0\%) & 53.3 (+10.2\%) \\
\bottomrule
\end{tabular*}
\end{table}

\subsection{Multi-Resolution Residual-Correction Forecaster Performance}\label{subsec:results_mrrc_forecaster}

Table~\ref{tab:mrrc_forecaster_comparison} compares Base, HFHR and MRRC forecasts over the complete 2021--2024 evaluation period for both TSFM backbones.

\begin{table}[!htbp]
\caption{Munich Base, HFHR and MRRC forecast performance pooled over 2021--2024. Brackets denote 95\% percentile confidence intervals.}\label{tab:mrrc_forecaster_comparison}
\small
\setlength{\tabcolsep}{4.5pt}
\begin{tabular*}{\tblwidth}{@{\extracolsep{\fill}}LRRRRRR@{}}
\toprule
& \multicolumn{3}{c}{Short-term forecast} & \multicolumn{2}{c}{12-hour integrated heat demand} & Computation \\
\cmidrule(lr){2-4}\cmidrule(lr){5-6}\cmidrule(l){7-7}
Setup & CVRMSE (\%) & $R^2$ & MAE (kW) & E-CVRMSE (\%) & E-bias (\%) & RTF \\
\midrule
\multicolumn{7}{l}{TabPFN-TS} \\
Base & \centeredtablecell{21.63\\{[19.95, 23.47]}} & \centeredtablecell{0.897\\{[0.880, 0.911]}} & \centeredtablecell{125.3\\{[119.0, 132.1]}} & \centeredtablecell{8.43\\{[7.79, 9.12]}} & \centeredtablecell{-0.99\\{[-1.38, -0.60]}} & $2.24{\times}10^{-5}$ \\
HFHR & \centeredtablecell{21.13\\{[19.37, 23.08]}} & \centeredtablecell{0.902\\{[0.884, 0.916]}} & \centeredtablecell{115.4\\{[109.6, 121.7]}} & \centeredtablecell{8.78\\{[7.92, 9.74]}} & \centeredtablecell{-1.77\\{[-2.25, -1.32]}} & $3.86{\times}10^{-4}$ \\
MRRC & \centeredtablecell{21.15\\{[19.38, 23.10]}} & \centeredtablecell{0.902\\{[0.884, 0.916]}} & \centeredtablecell{114.8\\{[109.1, 120.9]}} & \centeredtablecell{8.14\\{[7.52, 8.82]}} & \centeredtablecell{-1.09\\{[-1.44, -0.74]}} & $2.70{\times}10^{-4}$ \\
\midrule
\multicolumn{7}{l}{Chronos-2} \\
Base & \centeredtablecell{21.24\\{[19.55, 23.09]}} & \centeredtablecell{0.901\\{[0.884, 0.914]}} & \centeredtablecell{121.9\\{[115.6, 128.5]}} & \centeredtablecell{7.88\\{[7.22, 8.61]}} & \centeredtablecell{-0.66\\{[-0.97, -0.35]}} & $5.52{\times}10^{-7}$ \\
HFHR & \centeredtablecell{20.04\\{[18.34, 21.87]}} & \centeredtablecell{0.912\\{[0.896, 0.925]}} & \centeredtablecell{106.3\\{[101.1, 111.9]}} & \centeredtablecell{8.23\\{[7.33, 9.22]}} & \centeredtablecell{-0.18\\{[-0.47, 0.10]}} & $7.72{\times}10^{-6}$ \\
MRRC & \centeredtablecell{20.21\\{[18.44, 22.14]}} & \centeredtablecell{0.910\\{[0.893, 0.924]}} & \centeredtablecell{106.9\\{[101.7, 112.6]}} & \centeredtablecell{7.61\\{[6.97, 8.33]}} & \centeredtablecell{-0.66\\{[-0.94, -0.38]}} & $6.30{\times}10^{-6}$ \\
\bottomrule
\end{tabular*}
\end{table}

Relative to the Base Forecaster, MRRC significantly reduces short-term CVRMSE and MAE, increases $R^2$, and reduces 12-hour E-CVRMSE for both backbones. The relative short-term CVRMSE reduction is larger for Chronos-2, at 4.9\%, compared with 2.2\% for TabPFN-TS. The relative reduction in 12-hour E-CVRMSE is approximately 3.4\% for both.

Compared with direct HFHR forecasting, MRRC significantly reduces 12-hour E-CVRMSE by 7.3\% for TabPFN-TS and 7.5\% for Chronos-2. Its short-term benefits are less consistent: CVRMSE and $R^2$ differences are not statistically significant for either backbone, while MAE decreases slightly for TabPFN-TS and increases slightly for Chronos-2. Both MAE differences are statistically significant but below 1\% in relative terms. Paired confidence intervals are reported in Supplementary Table~S10.

MRRC also has lower recorded computational cost than HFHR, reducing RTF by 30\% for TabPFN-TS and 18\% for Chronos-2. The Base Forecaster remains the least expensive setup; MRRC requires approximately 12 and 11 times its recorded prediction time, respectively.

\section{Discussion}\label{sec:discussion}

Chronos-2 achieved stronger full-year deterministic performance and lower CRPS in both networks, consistent with the benefits of its time-series-specific architecture and pretraining. TabPFN-TS nevertheless remained within the critical-difference threshold of Chronos-2 in the daily CVRMSE-rank comparison in both networks. These results show that a model pretrained entirely on synthetic non-temporal tabular data can support competitive heat-load forecasting when combined with time-series featurization. Together with the solar-energy results of Liu et al. \cite{liu2026tabpfnsolar}, this supports the applicability of synthetically pretrained TabPFN models across different energy-related tasks. Task- and domain-tailored priors may offer further improvements.

The context-length diagnostics in Section~\ref{subsec:results_context_resolution_horizon} show that forecast accuracy does not improve monotonically with context length: longer histories can yield poorer forecasts than shorter ones. Optimal context length therefore remains application-specific even for foundation models. This may reflect superimposed demand processes: space heating is strongly seasonal and weather-driven, whereas domestic hot water is more stochastic and exhibits different, weaker temporal patterns. Without explicit separation, long contexts may include regimes irrelevant to the current forecast and obscure rather than improve in-context prediction.

The results support the ability of both TSFMs to capture predictable components of higher-frequency heat-load dynamics at 15-minute resolution. In the four-year comparison, the direct HFHR setup achieved lower short-term CVRMSE than the interpolated hourly Base Forecaster for both backbones in every year. Nevertheless, the configuration experiments showed lower forecasting errors at hourly resolution. This is likely due to temporal aggregation: hourly averaging smooths high-frequency dynamics that are difficult to predict at 15-minute resolution. One potential driver of these dynamics is domestic-hot-water draw-off behavior. The exact timing of individual draw-off events contains a stochastic component and is therefore inherently difficult to forecast \cite{fuentesReviewDomesticHot2018a}, whereas broader temporal patterns in domestic-hot-water demand are easier to capture \cite{maltaisPredictabilityAnalysisDomestic2021}. Additional sources of high-frequency variability may include hydraulic interactions between buildings \cite{xuFieldInvestigationConsumer2009}, varying transport delays between the energy hub and consumers, delay-dependent heat losses, hysteresis-controlled pressure pumps, and other operational effects. The short-horizon experiments also revealed a setting in which TabPFN-TS outperformed Chronos-2 in deterministic accuracy: for the 4-hour forecast task at 15-minute resolution, TabPFN-TS performed better than Chronos-2 when only very short context windows of 4 or 12 hours were available, as shown in Figure~\ref{fig:cvrmse_context_4h_quarter}. This is consistent with energy-demand forecasting results by Kim et al. \cite{kim2025dataefficient}, who report strong TabPFN performance under short-context conditions, and with the broader finding that TabPFN can deliver accurate predictions under low data availability \cite{hollmann2025accurate}.

Moving from daily day-ahead forecasts to weekly week-ahead forecasts reduced accuracy for both models, highlighting the importance of matching forecast horizon, update frequency and context length to the operational task.

The seasonal effects in Section~\ref{sec:seasonal_effects} provide complementary evidence on how demand composition affects forecast accuracy and uncertainty. The lower winter daily CVRMSE in Figure~\ref{fig:daily_weekly_horizon_metrics} is consistent with a larger share of weather-driven space-heating demand, creating a stronger and more predictable baseload. In summer, the smaller space-heating share and greater relative contribution of stochastic domestic hot water demand are expected to increase the relative forecast error. The higher \(R^2\) during the transitional period results from greater variation in measured demand, allowing more variance to be explained despite the relative error not being lowest. The greater accuracy loss from daily to weekly forecasts in this period may arise because the longer horizon spans rapidly changing load and weather conditions, making multi-day extrapolation harder than in the more stable winter and summer periods. The relative interval widths in Table~\ref{tab:prediction_interval_quality} similarly suggest that relative uncertainty rises as space-heating demand becomes less dominant and stochastic components gain importance. The transitional-week underprediction reported in Section~\ref{sec:seasonal_effects}, particularly for Chronos-2, may reflect difficulty adapting to changing operating conditions. One possible explanation is that the 12-week context window contains observations less representative of this regime.

The small, model-dependent improvements from additional weather covariates support ambient temperature as a parsimonious default for the evaluated forecasting task. This preference for a compact input set is consistent with Petkovi\'{c} et al. \cite{Petkovic2015InfluentialParametersHeatLoad}, who favor compact weather-input sets over large sets of meteorological covariates for heat load prediction. Because the additional computational cost was small in the tested setting, including further reliably available weather covariates remains defensible when sufficiently accurate future values are available.

The transfer results in Section~\ref{subsec:results_full_year_transfer} and Table~\ref{tab:full_year_transfer_benchmark} should be interpreted in light of network scale. The lower relative errors in the Flensburg validation case may partly reflect aggregation: domestic hot water demand is comparatively stochastic at the individual-building level but becomes more predictable as more buildings are connected. Because Flensburg is a city-scale district-heating system whereas Munich is a smaller mixed-use district, stronger aggregation in Flensburg may smooth idiosyncratic consumption events and contribute to its lower relative forecasting errors.

Our literature review in Section~\ref{subsec:heat_load_forecasting} identified retraining under concept drift as a central challenge in heat load forecasting. The four-year comparison in Section~\ref{subsec:results_drift_retraining} shows that the performance gap between static trained baselines and TSFMs widened over time, whereas the gap to baselines undergoing KSWIN-triggered batch retraining remained comparatively stable. This pattern supports our hypothesis that rolling-context in-context learning can accommodate changing system dynamics without task-specific weight updates. The remaining advantage over retrained baselines may reflect inherent model differences, the shorter 12-week context compared with a full training year that may retain more pre-change observations, and the TSFMs' ability to identify relevant dynamics within that context. Our experiments do not isolate the relative contributions of these factors.

As described in Section~\ref{subsec:results_context_resolution_horizon}, adding calendar-aligned observations from the previous year did not improve forecasts. However, the 2023 and 2024 data span the ongoing expansion of the Munich district and associated demand growth. The added block may therefore represent a different system configuration. Calendar alignment also only approximates physical similarity; selecting context by ambient temperature or clustered operating conditions may be more meaningful. However, even temperature-matched historical observations may be less representative if the network configuration or operating conditions have changed. Seasonally relevant context may thus be more beneficial in stationary district-heating systems.

Using coherent ECMWF temperature hindcasts increased CVRMSE by 7.4\% for TabPFN-TS and 10.1\% for Chronos-2 relative to the perfect-weather reference, while CRPS increased by 8.1\% and 10.2\%, respectively. These moderate relative increases suggest that both models retain much of their predictive accuracy under imperfect weather inputs in the evaluated setting. The extent of deterioration will nevertheless depend on the quality of the weather predictions available in operation.

For both backbones, TSFM-based residual correction significantly improved short-term CVRMSE, MAE and 12-hour E-CVRMSE relative to the uncorrected Base Forecaster over 2021--2024. This demonstrates the feasibility of using TSFMs as residual forecasters for TSFM predictions in the evaluated multi-resolution setup. Compared with HFHR, MRRC improved predictions of integrated heat demand over 12 hours while achieving broadly similar short-term accuracy. One possible explanation is its separation of forecasting time scales: the hourly Base Forecaster potentially captures slower variations associated with space heating, consistent with the dominant low-frequency patterns identified in the spectral analysis, while the Residual Forecaster addresses faster dynamics, including those associated with domestic hot water. Its computational savings arise from issuing expensive long-context base forecasts less frequently and using shorter-context residual forecasts for frequent updates. The MRRC architecture's benefit therefore depends on whether operational priorities favor short-term trajectory accuracy, longer-horizon heat-demand planning, or lower computational cost.

\subsection{Limitations}\label{subsec:limitations}

Although the two case studies cover DHNs with substantially different scales and building stocks, both are located in Germany and are therefore subject to broadly similar seasonal weather effects. Additionally, configuration selection and MRRC evaluation are restricted to Munich only. Generalization beyond these case studies therefore remains to be established. 

Configuration choices were evaluated on selected operating weeks rather than exhaustively optimized for every task. Results can depend on the evaluation period: TabPFN-TS's lower week-ahead CVRMSE on the selected operating weeks did not persist in the full-year evaluation. Findings from selected weeks therefore need not generalize to year-round performance. 

The principal benchmark assumes perfect weather forecasts by using realized future temperatures. The weather sensitivity analysis uses retrospective hindcasts rather than verified forecasts delivered operationally in 2024. The sensitivity analysis of weather forecast uncertainty covers only 24-hour heat-load predictions. Its findings therefore cannot be directly extended to longer forecasting horizons, for which increasing weather forecast uncertainty may have a larger effect on heat-load prediction.

KSWIN-triggered batch retraining uses the preceding year of observations and trains the models from scratch. Consequently, observations preceding a detected change can remain in the training window; no explicit change-based selection of training data is applied.

Finally, the HFHR setup used as a benchmark in the MRRC experiments was not independently optimized for its 13-hour forecast horizon and increased update frequency.

\section{Conclusion \& Future Work}\label{sec:conclusion}

This work systematically evaluated TabPFN-TS and Chronos-2 for zero-shot probabilistic heat load forecasting in DHNs. The results demonstrate the suitability of both evaluated foundation models for this task, with Chronos-2 achieving the strongest overall forecast quality and TabPFN-TS also proving competitive. For hourly 24-hour forecasting, both models achieved their lowest error among the tested contexts with a 12-week window, whereas shorter contexts performed better in some high-resolution settings. Ambient temperature alone provided an effective weather-covariate set.

The experiments show that rolling-context foundation-model forecasting can perform well in an evolving DHN without repeated task-specific retraining. Both foundation models performed strongly in the Flensburg validation case and outperformed the trained AutoGluon baselines in the main full-year comparison. At the same time, the diagnostic experiments showed that configuration choices remain important: more context data, higher temporal resolution, and larger weather-covariate sets did not automatically improve performance.

The Multi-Resolution Residual-Correction Forecaster illustrates how residual correction can be adapted to DHN operational requirements. Compared with direct high-frequency, high-resolution forecasting, MRRC significantly reduced 12-hour E-CVRMSE for both backbones and required less recorded prediction time, while short-term accuracy remained broadly similar.

Future work on MRRC should address both probabilistic forecasting and further configuration and architecture development. A probabilistic extension requires representing uncertainty consistently across the hourly base forecast and quarter-hourly residual correction, including beyond the two-hour residual horizon. Additionally, configuration and architecture studies could vary the Residual Forecaster horizon, test multiple residual-correction stages, and compare combinations of TSFMs and classical machine-learning models as Base and Residual Forecasters. The separation of slow load trends and high-frequency deviations could also be evaluated in other energy-system forecasting tasks.

A second direction for future work is to adapt foundation models more closely to district-heating heat load forecasting by preconditioning them on real-world time series, simulation data, or synthetic data containing the relevant dynamics. Task-specific prior data or further weight adaptation may improve forecast accuracy. However, if target-network data are included in the prior or used for adaptation, robustness under concept drift must be evaluated separately. For the trained baselines, future work could investigate whether excluding pre-change training data or updating existing models through warm-starting or transfer learning improves adaptation and reduces retraining costs.

Finally, future work should extend forecasts beyond heat demand to mass flow rates as well as supply and return temperatures. The delivered heat rate is proportional to mass flow rate and temperature difference, while the efficiency and usability of sustainable sources such as heat pumps and waste-heat streams depend on the required temperature level. Spatially distributed multivariate forecasts within the network could further support decentralized heat-source integration by providing more detailed information for operational planning and control.

\section*{CRediT Authorship Contribution Statement}

\noindent\textbf{Ben Spoek:} Conceptualization, Methodology, Software, Validation, Formal analysis, Investigation, Data curation, Visualization, Writing -- original draft, Project administration.

\noindent\textbf{Karim K. {Ben Hicham}:} Conceptualization, Methodology, Software, Validation, Writing -- review and editing.

\noindent\textbf{Kai Derzsi:} Data curation, Methodology, Validation, Formal analysis, Writing -- review and editing.

\noindent\textbf{Philipp Althaus:} Conceptualization, Supervision, Writing -- review and editing.

\noindent\textbf{Alexander Mitsos:} Supervision, Project administration, Funding acquisition, Writing -- review and editing.

\noindent\textbf{Dirk M{\"u}ller:} Supervision, Project administration, Funding acquisition, Writing -- review and editing.

\section*{Funding}

We gratefully acknowledge the financial support provided by the Federal Ministry for Economic Affairs and Energy (BMWE) under funding code 03EN3118.
We acknowledge support of the Werner Siemens Foundation in the frame of the WSS Research Center
``catalaix''.
The funders had no role in the study design, data collection, analysis, interpretation, manuscript preparation, or decision to submit the article.

\section*{Declaration of Competing Interest}

The authors declare that they have no known competing financial interests or personal relationships that could have appeared to influence the work reported in this paper.

\section*{Code \& Data Availability}

The code used for the experiments is available on GitHub at \url{https://github.com/benspoek/tsfm-heatload} and on Zenodo at \url{https://doi.org/10.5281/zenodo.23101871}. The Munich district heating data set used in this study is not publicly available due to third-party restrictions on the underlying operational data. The Flensburg district heating data set is a third-party data set and is publicly available.\cite{freissmannNetworkDataDistrict2025}

\section*{Acknowledgments and Computational Resources}

Computations were performed with computing resources granted by RWTH Aachen University under project rwth2083.

\section*{Declaration of Generative AI and AI-Assisted Technologies in the Manuscript Preparation Process}

During the preparation of this work, the authors used OpenAI Codex to assist with code review and editing, manuscript organization, language refinement, and consistency checks. The authors reviewed and edited the output as needed and take full responsibility for the content of the published article.

\bibliographystyle{vancouver}
\bibliography{references}

\clearpage
\appendix
\setcounter{section}{0}
\setcounter{figure}{0}
\setcounter{table}{0}
\setcounter{equation}{0}
\renewcommand{\thesection}{S\arabic{section}}
\renewcommand{\thefigure}{S\arabic{figure}}
\renewcommand{\thetable}{S\arabic{table}}
\renewcommand{\theequation}{S\arabic{equation}}
\renewcommand{\theHsection}{S.\arabic{section}}
\renewcommand{\theHfigure}{S.\arabic{figure}}
\renewcommand{\theHtable}{S.\arabic{table}}
\renewcommand{\theHequation}{S.\arabic{equation}}

\section*{Supplementary Information}
\addcontentsline{toc}{section}{Supplementary Information}
\captionsetup[table]{font={small,sf},labelfont=bf,labelsep=newline,
justification=raggedright,singlelinecheck=false,position=top,skip=7pt}
\input{supplementary}

\end{document}

%% file: supplementary.tex
This Supplementary Information provides network and operating-week details, weather-covariate configurations, additional forecasting and sensitivity results, and computational costs.

\paragraph{Shared statistical procedure.}
Throughout the supplementary tables, brackets denote pointwise 95\% percentile confidence intervals computed from 10,000 paired bootstrap samples. Tables~\ref{tab:supp_weather_covariate_configurations}, \ref{tab:supp_anomaly_benchmark}--\ref{tab:supp_kswin_annual_metrics} and~\ref{tab:supp_mrrc_paired} use circular moving blocks of seven consecutive forecast days. Identical blocks are used across compared models or configurations, and aggregate metrics are recomputed in each sample. Table~\ref{tab:supp_full_year_weekly_horizon} resamples complete weekly forecasts, as described in Section~\ref{sec:supp_full_year_weekly_horizon}.

\section{Munich Network Development}

\begin{figure}[H]
  \centering
  \includegraphics[width=0.78\textwidth]{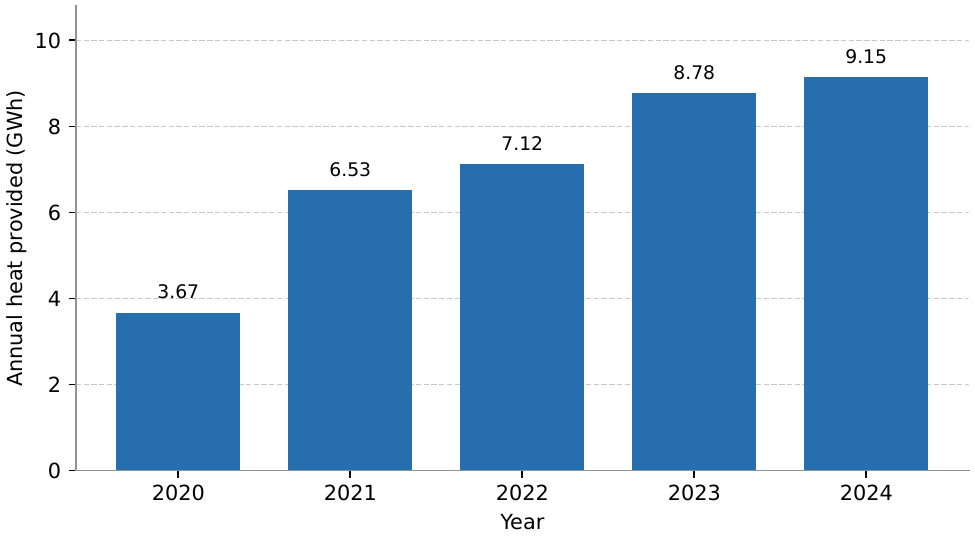}
  \caption{Annual heat provided by the Munich district heating network for complete calendar years.}
  \label{fig:supp_munich_yearly_heat_gwh}
\end{figure}

Annual heat supply documents the development of the Munich network over the available observation period. Figure~\ref{fig:supp_munich_yearly_heat_gwh} includes each complete calendar year. Annual heat supply increased as the network expanded.

\section{Selection of Representative Operating Weeks}

Representative weeks provide contrasting operating conditions for the configuration diagnostics. We selected complete Monday--Sunday weeks without observations flagged by the screening procedure described in Section~3.3 of the main manuscript. From the remaining weeks, the coldest week represents winter, the hottest week represents summer, and the week with the largest within-week ambient-temperature variability represents transitional operation. Table~\ref{tab:supp_selected_weeks_temperature} reports the selected weeks and their temperature statistics.

\begin{table}[H]
\sitableformat
\caption{Selected representative weeks from 2024 and their ambient- and supply-temperature statistics. Supply temperatures are reported as mean $\pm$ standard deviation (SD) of the 672 unfiltered 15-minute observations in each week.}
\label{tab:supp_selected_weeks_temperature}
\begin{tabular*}{\linewidth}{@{\extracolsep{\fill}}lrrrcl@{}}
\toprule
& \multicolumn{3}{c}{Ambient temperature ($^\circ$C)} & Supply ($^\circ$C) & \\
\cmidrule(lr){2-4}
Regime & Min. & Mean & Max. & Mean $\pm$ SD & Dates \\
\midrule
Winter & -8.60 & -0.51 & 11.20 & $86.42 \pm 4.42$ & Jan. 15--21 \\
Transitional & 0.40 & 7.97 & 24.60 & $83.63 \pm 4.12$ & Apr. 22--28 \\
Summer & 16.80 & 23.16 & 32.20 & $84.14 \pm 3.81$ & Aug. 12--18 \\
\bottomrule
\end{tabular*}
\end{table}

\section{Weather-Forecast Configurations and Sensitivity}\label{sec:supp_weather_covariates}

This comparison examines how measured and forecast temperature histories affect forecasts and quantifies the change relative to the perfect-weather reference. We compared three ways to incorporate measured and forecast temperatures into each time-series foundation model. In Table~\ref{tab:supp_weather_covariate_configurations}, measured and forecast temperatures are denoted by M and F, respectively. The first configuration uses forecast temperatures in both the historical context and future horizon. The second uses a single covariate column containing historical measurements and future forecasts. The third uses separate measured- and forecast-temperature columns in the historical context. Chronos-2 receives only forecast temperature over the future horizon. Because the TabPFN-TS implementation requires future values for both columns, both contain the same forecast temperature over the future horizon.

All configurations use the same Munich target observations, 12-week context, hourly resolution and 24-hour horizon, with 208 daily forecast starts from 7 June to 31 December 2024, comprising 4,992 hourly predictions. The forecast-weather configurations use coherent retrospective ECMWF temperature predictions. The M$\rightarrow$M configuration corresponds to our perfect-weather forecast baseline, using measured temperatures throughout, including the future horizon. For Table~\ref{tab:supp_weather_covariate_configurations}, each resampled day retains all 24 horizon observations.

\begin{table}[h]
\sitableformat
\caption{Temperature-covariate configurations on matched Munich forecast starts. The left and right sides of each arrow specify historical and future covariates.}
\label{tab:supp_weather_covariate_configurations}
\setlength{\tabcolsep}{3pt}
\begin{tabular*}{\linewidth}{@{\extracolsep{\fill}}llrrrr@{}}
\toprule
Model & Configuration & CVRMSE (\%) & $R^2$ & MAE (kW) & CRPS (kW) \\
\midrule
TabPFN-TS & M$\rightarrow$M & \centeredtablecell{11.89\\{[9.93, 14.46]}} & \centeredtablecell{0.974\\{[0.961, 0.982]}} & \centeredtablecell{72.2\\{[61.7, 83.0]}} & \centeredtablecell{50.85\\{[43.22, 58.62]}} \\
\addlinespace[3pt]
 & F$\rightarrow$F & \centeredtablecell{14.85\\{[12.87, 17.12]}} & \centeredtablecell{0.960\\{[0.941, 0.971]}} & \centeredtablecell{92.0\\{[75.3, 109.8]}} & \centeredtablecell{64.16\\{[52.57, 76.38]}} \\
\addlinespace[3pt]
 & M$\rightarrow$F & \centeredtablecell{13.00\\{[11.12, 15.36]}} & \centeredtablecell{0.969\\{[0.955, 0.978]}} & \centeredtablecell{78.9\\{[66.7, 91.7]}} & \centeredtablecell{55.96\\{[47.09, 65.16]}} \\
\addlinespace[3pt]
 & M+F$\rightarrow$F+F & \centeredtablecell{12.77\\{[10.91, 15.13]}} & \centeredtablecell{0.971\\{[0.957, 0.979]}} & \centeredtablecell{77.8\\{[66.1, 89.9]}} & \centeredtablecell{54.95\\{[46.52, 63.67]}} \\
\midrule
Chronos-2 & M$\rightarrow$M & \centeredtablecell{11.41\\{[9.34, 14.10]}} & \centeredtablecell{0.976\\{[0.963, 0.984]}} & \centeredtablecell{68.5\\{[58.9, 78.4]}} & \centeredtablecell{48.32\\{[41.31, 55.47]}} \\
\addlinespace[3pt]
 & F$\rightarrow$F & \centeredtablecell{12.74\\{[10.83, 15.18]}} & \centeredtablecell{0.971\\{[0.956, 0.979]}} & \centeredtablecell{77.0\\{[65.0, 89.6]}} & \centeredtablecell{53.85\\{[45.20, 62.79]}} \\
\addlinespace[3pt]
 & M$\rightarrow$F & \centeredtablecell{12.94\\{[11.05, 15.36]}} & \centeredtablecell{0.970\\{[0.956, 0.978]}} & \centeredtablecell{77.9\\{[66.2, 90.2]}} & \centeredtablecell{55.22\\{[46.57, 64.27]}} \\
\addlinespace[3pt]
 & M+F$\rightarrow$F & \centeredtablecell{12.56\\{[10.63, 15.04]}} & \centeredtablecell{0.971\\{[0.957, 0.980]}} & \centeredtablecell{76.1\\{[64.3, 88.2]}} & \centeredtablecell{53.27\\{[44.89, 61.86]}} \\
\bottomrule
\end{tabular*}
\end{table}

Separate measured- and forecast-temperature histories give the lowest CVRMSE, MAE and CRPS among the tested forecast-weather configurations for both models. The differences between configurations are larger for TabPFN-TS than for Chronos-2. These configurations were selected for the main sensitivity comparison. Compared with the perfect-weather reference, the selected configurations increase CRPS by 4.10~kW for TabPFN-TS and 4.95~kW for Chronos-2, with paired 95\% confidence intervals of 2.01--6.38~kW and 2.76--7.30~kW, respectively.

Paired differences between the selected forecast-weather configurations and the perfect-weather reference use the same 208 forecast starts and the shared bootstrap procedure. Forecast weather increases CVRMSE and MAE and reduces \(R^2\) for both models.

\section{Full-Year Week-Ahead Forecasting}\label{sec:supp_full_year_weekly_horizon}

On the selected operating weeks, TabPFN-TS achieved lower week-ahead CVRMSE than Chronos-2, reversing the day-ahead ranking reported in main-text Section~5.1 and Figure~5. We therefore repeated the week-ahead evaluation over the full year to assess whether this advantage was specific to the selected weeks or persisted across the year. Table~\ref{tab:supp_full_year_weekly_horizon} reports the results. Confidence intervals are computed from 10,000 paired bootstrap samples, each containing 52 weekly forecasts drawn with replacement. Each selected forecast retains all 168 hourly predictions. In contrast to the selected-week CVRMSE results, Chronos-2 achieves lower CVRMSE and MAE and higher \(R^2\) than TabPFN-TS.

\begin{table}[H]
\sitableformat
\caption{Full-year Munich results for non-overlapping weekly 168-hour forecasts at hourly resolution.}
\label{tab:supp_full_year_weekly_horizon}
\begin{tabular*}{\linewidth}{@{\extracolsep{\fill}}lrrr@{}}
\toprule
Model & CVRMSE (\%) & \(R^2\) & MAE (kW) \\
\midrule
Chronos-2 & 15.64 [13.44, 18.05] & 0.941 [0.914, 0.958] & 109.7 [93.8, 126.9] \\
TabPFN-TS & 17.06 [14.58, 19.94] & 0.929 [0.895, 0.950] & 120.2 [102.2, 140.1] \\
\bottomrule
\end{tabular*}
\end{table}

\section{Sensitivity to Anomaly Exclusion}\label{sec:supp_anomaly_sensitivity}

This analysis assesses whether potentially anomalous evaluation targets affect the full-year benchmark conclusions. Excluding flagged Munich days leaves 342 of 366 days; no Flensburg day met the screening criteria. Table~\ref{tab:supp_anomaly_benchmark} reports absolute filtered scores in Panel A and paired filtered-minus-unfiltered changes for all evaluated models in Panel B.

\begin{table}[h]
\sitableformat
\caption{Munich anomaly-exclusion sensitivity, retaining 342 of 366 forecast days. pp denotes percentage points.}
\label{tab:supp_anomaly_benchmark}
\setlength{\tabcolsep}{4pt}
\begin{tabular*}{\linewidth}{@{\extracolsep{\fill}}lrrr@{}}
\toprule
\multicolumn{4}{c}{\textbf{A. Scores after anomaly exclusion}} \\
\midrule
Model & CVRMSE (\%) & $R^2$ & MAE (kW) \\
\midrule
Chronos-2 & 11.83 [10.27, 13.76] & 0.966 [0.950, 0.975] & 82.1 [72.7, 92.2] \\
TabPFN-TS & 12.58 [11.02, 14.49] & 0.961 [0.944, 0.971] & 87.8 [77.2, 98.8] \\
TFT & 14.24 [12.62, 16.20] & 0.950 [0.930, 0.962] & 99.7 [88.1, 112.1] \\
Direct LGBM & 15.47 [13.81, 17.40] & 0.941 [0.919, 0.955] & 110.5 [95.8, 126.4] \\
DeepAR & 16.09 [14.46, 17.96] & 0.936 [0.914, 0.952] & 113.8 [98.0, 130.6] \\
AutoGluon Ensemble & 16.09 [14.46, 17.96] & 0.936 [0.914, 0.952] & 113.8 [98.0, 130.6] \\
Direct XGBoost & 16.82 [15.07, 18.78] & 0.931 [0.907, 0.946] & 122.6 [106.9, 140.2] \\
ETS & 18.09 [16.06, 20.44] & 0.920 [0.887, 0.940] & 129.6 [113.3, 146.2] \\
AutoETS & 18.09 [16.06, 20.44] & 0.920 [0.887, 0.940] & 129.6 [113.3, 146.2] \\
DLinear & 18.46 [16.44, 20.81] & 0.916 [0.883, 0.937] & 133.0 [116.0, 150.4] \\
Feed-forward & 19.24 [17.20, 21.62] & 0.909 [0.874, 0.931] & 139.9 [121.9, 158.2] \\
AutoARIMA & 20.34 [18.04, 22.99] & 0.898 [0.857, 0.924] & 146.2 [126.9, 165.6] \\
Recursive LGBM & 21.03 [18.67, 23.84] & 0.891 [0.848, 0.919] & 150.9 [130.9, 171.3] \\
Seasonal naive & 22.46 [19.90, 25.46] & 0.876 [0.826, 0.908] & 160.7 [139.0, 182.7] \\
Naive & 23.86 [20.84, 27.51] & 0.860 [0.796, 0.899] & 173.0 [152.3, 195.3] \\
\bottomrule
\end{tabular*}
\par\medskip
\begin{tabular*}{\linewidth}{@{\extracolsep{\fill}}lrrr@{}}
\toprule
\multicolumn{4}{c}{\textbf{B. Paired differences: anomaly-filtered minus unfiltered}} \\
\midrule
Model & $\Delta$CVRMSE (pp) & $\Delta R^2$ & $\Delta$MAE (kW) \\
\midrule
Chronos-2 & $-0.65$ [$-1.61$, $-0.03$] & 0.0034 [$-0.0006$, 0.0098] & $-3.28$ [$-5.84$, $-1.07$] \\
TabPFN-TS & $-0.49$ [$-1.38$, 0.07] & 0.0026 [$-0.0015$, 0.0090] & $-2.47$ [$-4.84$, $-0.40$] \\
TFT & $-0.37$ [$-1.08$, 0.14] & 0.0020 [$-0.0023$, 0.0081] & $-2.41$ [$-4.57$, $-0.48$] \\
Direct LGBM & $-0.40$ [$-1.09$, 0.08] & 0.0023 [$-0.0019$, 0.0083] & $-3.19$ [$-5.89$, $-1.00$] \\
DeepAR & $-0.45$ [$-1.25$, 0.10] & 0.0028 [$-0.0028$, 0.0106] & $-3.38$ [$-6.89$, $-0.50$] \\
AutoGluon Ensemble & $-0.45$ [$-1.25$, 0.10] & 0.0028 [$-0.0028$, 0.0106] & $-3.38$ [$-6.89$, $-0.50$] \\
Direct XGBoost & $-0.52$ [$-1.16$, 0.005] & 0.0035 [$-0.0008$, 0.0094] & $-4.04$ [$-7.61$, $-1.29$] \\
ETS & $-0.34$ [$-1.04$, 0.21] & 0.0021 [$-0.0038$, 0.0096] & $-3.23$ [$-6.07$, $-0.80$] \\
AutoETS & $-0.34$ [$-1.04$, 0.21] & 0.0021 [$-0.0038$, 0.0096] & $-3.23$ [$-6.07$, $-0.80$] \\
DLinear & $-0.20$ [$-0.81$, 0.31] & 0.0008 [$-0.0054$, 0.0080] & $-2.08$ [$-4.92$, 0.65] \\
Feed-forward & $-0.16$ [$-0.74$, 0.32] & 0.0005 [$-0.0057$, 0.0075] & $-1.95$ [$-4.64$, 0.70] \\
AutoARIMA & $-0.14$ [$-0.77$, 0.40] & 0.0002 [$-0.0068$, 0.0078] & $-2.08$ [$-4.93$, 0.33] \\
Recursive LGBM & $-0.05$ [$-0.53$, 0.41] & $-0.0008$ [$-0.0074$, 0.0057] & $-1.09$ [$-3.53$, 1.62] \\
Seasonal naive & $-0.05$ [$-0.54$, 0.39] & $-0.0008$ [$-0.0079$, 0.0061] & $-1.29$ [$-4.09$, 1.73] \\
Naive & 0.20 [$-0.30$, 0.72] & $-0.0039$ [$-0.0129$, 0.0031] & 0.89 [$-2.04$, 4.51] \\
\bottomrule
\end{tabular*}
\end{table}

The CVRMSE ranking remains unchanged, with Chronos-2, TabPFN-TS and TFT retaining the leading positions. CVRMSE and MAE point estimates decrease for 14 of the 15 models; the naive baseline is the exception. Paired confidence intervals exclude zero for MAE reductions in nine models and for the CVRMSE reduction in Chronos-2 only. All intervals for changes in \(R^2\) include zero.

\clearpage
\section{Probabilistic Forecast Evaluation}\label{sec:supp_probabilistic_quality}

The full-year CRPS results complement the deterministic benchmark, while seasonal PIT histograms examine calibration across operating conditions. Table~\ref{tab:supp_full_year_crps} reports CRPS for both networks and for the Munich anomaly-exclusion sensitivity using the shared bootstrap procedure.

\begin{table}[H]
\sitableformat
\caption{Probabilistic forecast quality for the 2024 full-year benchmark and anomaly-exclusion sensitivity.}
\label{tab:supp_full_year_crps}
\setlength{\tabcolsep}{4pt}
\begin{tabular*}{\linewidth}{@{\extracolsep{\fill}}lllr@{}}
\toprule
Network & Anomaly exclusion & Model & CRPS (kW) \\
\midrule
Munich & No & TabPFN-TS & 63.66 [56.26, 71.39] \\
Munich & No & Chronos-2 & 60.49 [53.67, 67.83] \\
Munich & Yes & TabPFN-TS & 61.54 [54.08, 69.43] \\
Munich & Yes & Chronos-2 & 57.90 [51.11, 65.21] \\
Flensburg & No & TabPFN-TS & 5942 [5115, 6825] \\
Flensburg & No & Chronos-2 & 5618 [4821, 6484] \\
\bottomrule
\end{tabular*}
\end{table}

Chronos-2 achieves lower CRPS in both networks, and anomaly exclusion reduces CRPS for both models in Munich.

\begin{figure}[H]
\centering
\includegraphics[width=\linewidth,height=0.48\textheight,keepaspectratio]{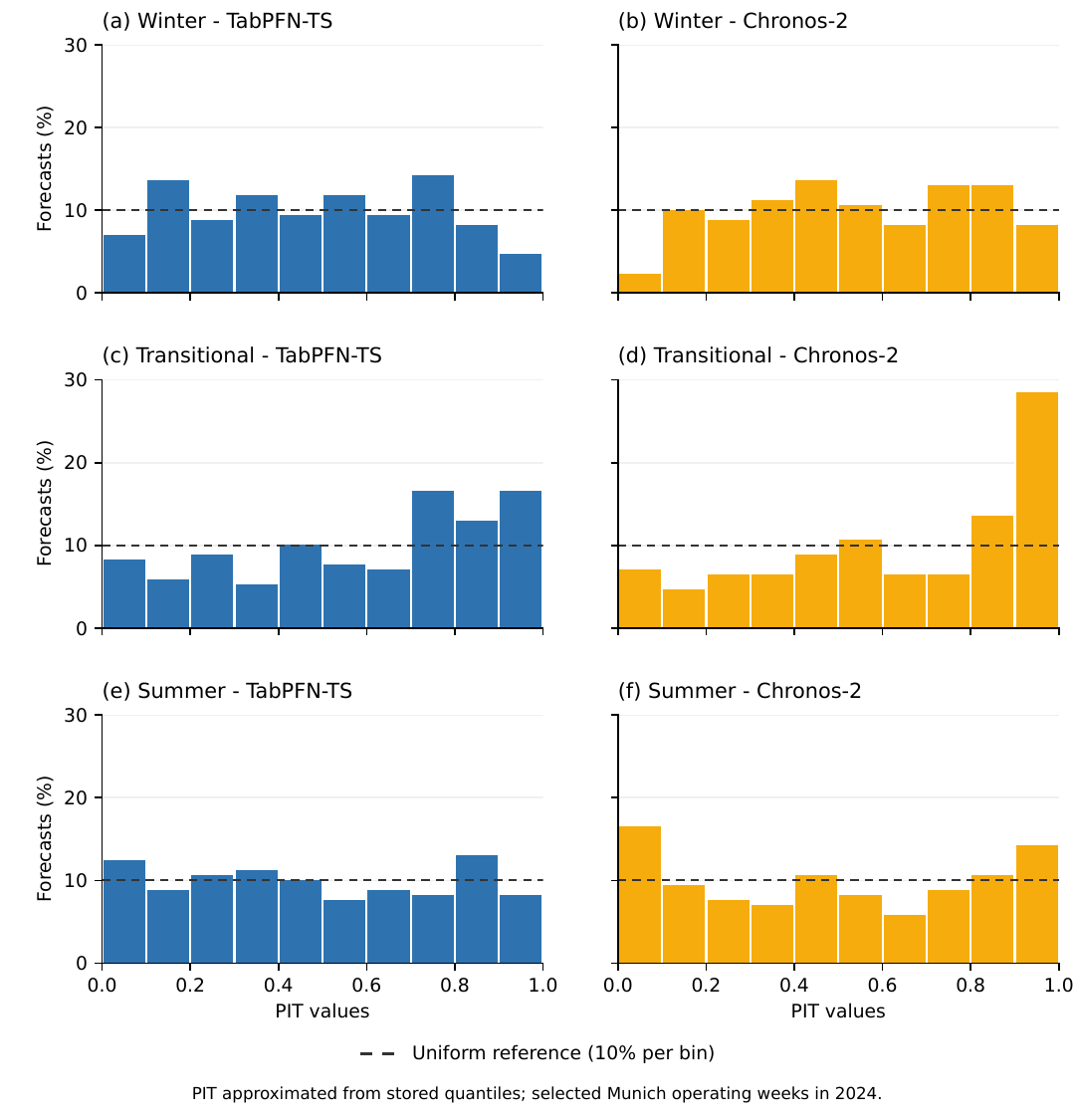}
\caption{PIT histograms for TabPFN-TS and Chronos-2 during the selected Munich operating weeks in 2024.}
\label{fig:supp_seasonal_pit}
\end{figure}

Figure~\ref{fig:supp_seasonal_pit} shows PIT histograms for the selected operating weeks.

\clearpage
\section{Baseline Retraining on Detected Concept Drift}\label{sec:supp_kswin_retraining}

This comparison assesses the effect of updating trained baselines during deployment. We compare Munich forecasts over 2021--2024 using AutoGluon models trained once on 2020 data or subsequently batch-retrained from scratch following KSWIN drift detections.

Table~\ref{tab:supp_kswin_annual_metrics} complements the CVRMSE comparison in main-text Table~7 with annual MAE and \(R^2\) results. Scores are averaged equally across TFT, DeepAR, DLinear, Feed-forward, Direct LGBM, Direct XGBoost, Recursive LGBM, and AutoGluon Ensemble. Differences are KSWIN minus static. KSWIN-triggered retraining reduces average MAE and increases average \(R^2\) in every year.

\begin{table}[h]
\sitableformat
\caption{Annual mean MAE and $R^2$ across eight learned AutoGluon setups in Munich.}
\label{tab:supp_kswin_annual_metrics}
\begin{tabular*}{\linewidth}{@{\extracolsep{\fill}}ccccc@{}}
\toprule
\multicolumn{5}{c}{\textbf{A. MAE}} \\
\midrule
Year & Static MAE (kW) & KSWIN MAE (kW) & $\Delta$MAE (kW) & \centeredtablecell{Relative\\reduction (\%)} \\
\midrule
2021 & \centeredtablecell{105.3\\{[95.8, 115.0]}} & \centeredtablecell{100.0\\{[92.6, 107.4]}} & \centeredtablecell{-5.3\\{[-9.3, -1.8]}} & 5.0 \\
2022 & \centeredtablecell{112.7\\{[101.7, 124.4]}} & \centeredtablecell{101.0\\{[92.8, 109.8]}} & \centeredtablecell{-11.7\\{[-17.6, -6.5]}} & 10.4 \\
2023 & \centeredtablecell{158.7\\{[137.4, 181.3]}} & \centeredtablecell{134.1\\{[119.1, 150.4]}} & \centeredtablecell{-24.6\\{[-33.9, -15.7]}} & 15.5 \\
2024 & \centeredtablecell{150.2\\{[131.5, 170.5]}} & \centeredtablecell{120.1\\{[106.3, 134.7]}} & \centeredtablecell{-30.1\\{[-40.1, -21.3]}} & 20.0 \\
\bottomrule
\end{tabular*}
\par\medskip
\begin{tabular*}{\linewidth}{@{\extracolsep{\fill}}ccccc@{}}
\toprule
\multicolumn{5}{c}{\textbf{B. $R^2$}} \\
\midrule
Year & Static $R^2$ & KSWIN $R^2$ & $\Delta R^2$ & \centeredtablecell{Relative\\increase (\%)} \\
\midrule
2021 & \centeredtablecell{0.872\\{[0.840, 0.892]}} & \centeredtablecell{0.885\\{[0.851, 0.906]}} & \centeredtablecell{0.0128\\{[0.0047, 0.0208]}} & 1.5 \\
2022 & \centeredtablecell{0.905\\{[0.875, 0.923]}} & \centeredtablecell{0.925\\{[0.891, 0.943]}} & \centeredtablecell{0.0195\\{[0.0106, 0.0286]}} & 2.1 \\
2023 & \centeredtablecell{0.872\\{[0.837, 0.895]}} & \centeredtablecell{0.909\\{[0.877, 0.929]}} & \centeredtablecell{0.0368\\{[0.0247, 0.0506]}} & 4.2 \\
2024 & \centeredtablecell{0.891\\{[0.863, 0.909]}} & \centeredtablecell{0.929\\{[0.904, 0.946]}} & \centeredtablecell{0.0390\\{[0.0265, 0.0537]}} & 4.4 \\
\bottomrule
\end{tabular*}
\end{table}

Table~\ref{tab:supp_kswin_retraining} shows how the aggregate refitting effects differ between models over the complete four-year period. A dash indicates methods for which the evaluated training and refitting strategy does not apply. Direct XGBoost and Direct LGBM show the largest CVRMSE reductions, whereas DLinear and Feed-forward deteriorate after retraining. Both TSFMs retain lower aggregate errors than the static and refitted baselines.

\begin{table}[h]
\sitableformat
\caption{Forecast performance over 2021--2024 in Munich.}
\label{tab:supp_kswin_retraining}
\setlength{\tabcolsep}{7pt}
\renewcommand{\arraystretch}{1.15}
\begin{tabular*}{\linewidth}{@{\extracolsep{\fill}}ccccc@{}}
\toprule
Model & \centeredtablecell{KSWIN triggered\\retraining} & CVRMSE (\%) & MAE (kW) & $R^2$ \\
\midrule
TabPFN-TS & -- & 14.83 & 86.1 & 0.949 \\
Chronos-2 & -- & 14.33 & 81.9 & 0.952 \\
\midrule
TFT & \centeredtablecell{No\\Yes} & \centeredtablecell{16.94\\16.57} & \centeredtablecell{104.8\\98.9} & \centeredtablecell{0.933\\0.936} \\
\addlinespace[3pt]
AutoGluon Ensemble & \centeredtablecell{No\\Yes} & \centeredtablecell{16.64\\16.26} & \centeredtablecell{102.3\\97.3} & \centeredtablecell{0.935\\0.938} \\
\addlinespace[3pt]
DeepAR & \centeredtablecell{No\\Yes} & \centeredtablecell{20.34\\16.38} & \centeredtablecell{131.2\\98.4} & \centeredtablecell{0.904\\0.937} \\
\addlinespace[3pt]
Direct LGBM & \centeredtablecell{No\\Yes} & \centeredtablecell{26.21\\18.81} & \centeredtablecell{164.1\\119.0} & \centeredtablecell{0.840\\0.918} \\
\addlinespace[3pt]
Direct XGBoost & \centeredtablecell{No\\Yes} & \centeredtablecell{30.03\\19.31} & \centeredtablecell{182.5\\122.7} & \centeredtablecell{0.790\\0.913} \\
\addlinespace[3pt]
Recursive LGBM & \centeredtablecell{No\\Yes} & \centeredtablecell{21.62\\20.75} & \centeredtablecell{132.9\\127.6} & \centeredtablecell{0.891\\0.900} \\
\addlinespace[3pt]
DLinear & \centeredtablecell{No\\Yes} & \centeredtablecell{18.99\\19.79} & \centeredtablecell{117.2\\122.3} & \centeredtablecell{0.916\\0.909} \\
\addlinespace[3pt]
Feed-forward & \centeredtablecell{No\\Yes} & \centeredtablecell{19.23\\19.97} & \centeredtablecell{118.9\\124.1} & \centeredtablecell{0.914\\0.907} \\
\midrule
AutoARIMA & -- & 20.62 & 129.0 & 0.901 \\
AutoETS & -- & 19.19 & 118.8 & 0.914 \\
ETS & -- & 19.19 & 118.8 & 0.914 \\
Naive & -- & 24.00 & 147.3 & 0.866 \\
Seasonal naive & -- & 23.42 & 144.1 & 0.872 \\
\bottomrule
\end{tabular*}
\end{table}

\clearpage
\section{Computational Costs of Trained Baselines}

Table~\ref{tab:supp_ag_initial_timing} reports marginal fitting times for the neural and tabular models from the controlled initial fit on 2023 data. The complete AutoGluon suite took 159.25~s, including validation and framework overhead.

\begin{table}[H]
\centering
\begin{minipage}{0.72\linewidth}
\sitableformat
\caption{Initial marginal AutoGluon fitting times on the 2023 training data.}
\label{tab:supp_ag_initial_timing}
\begin{tabular*}{\linewidth}{@{\extracolsep{\fill}}lc@{}}
\toprule
Model & \centeredtablecell{Initial marginal\\fitting time (s)} \\
\midrule
TFT & 47.613 \\
DeepAR & 31.920 \\
Direct LGBM & 0.729 \\
Direct XGBoost & 1.044 \\
Recursive LGBM & 0.572 \\
DLinear & 14.335 \\
Feed-forward & 8.644 \\
\bottomrule
\end{tabular*}
\end{minipage}
\end{table}

Table~\ref{tab:supp_kswin_fit_times} reports the fitting-to-prediction time ratio, calculated as cumulative marginal fitting time across the 149 KSWIN-triggered refits divided by cumulative prediction-call time for the corresponding model. A ratio of 1 means that refitting takes as much time as prediction.

\begin{table}[H]
\centering
\begin{minipage}{0.72\linewidth}
\sitableformat
\caption{Additional fitting cost of KSWIN-triggered retraining over 2021--2024.}
\label{tab:supp_kswin_fit_times}
\begin{tabular*}{\linewidth}{@{\extracolsep{\fill}}lr@{}}
\toprule
Model & \centeredtablecell{Fitting-to-prediction\\time ratio} \\
\midrule
TFT & 5.706 \\
DeepAR & 26.603 \\
Direct LGBM & 1.595 \\
Direct XGBoost & 0.491 \\
Recursive LGBM & 0.254 \\
DLinear & 45.761 \\
Feed-forward & 78.157 \\
\bottomrule
\end{tabular*}
\end{minipage}
\end{table}
\clearpage
\section{Multi-Year MRRC Evaluation}\label{sec:supp_mrrc}

Table~\ref{tab:supp_mrrc_paired} reports paired MRRC-minus-comparator differences over 2021--2024, supporting the comparisons in main-text Table~9.
\begin{table}[H]
\sitableformat
\caption{Paired MRRC-minus-comparator differences for the pooled 2021--2024 evaluation.}
\label{tab:supp_mrrc_paired}
\begin{tabular*}{\linewidth}{@{\extracolsep{\fill}}llccc@{}}
\toprule
Backbone & Comparison & \centeredtablecell{$\Delta$CVRMSE\\(pp)} & \centeredtablecell{$\Delta$MAE\\(kW)} & \centeredtablecell{$\Delta$E-CVRMSE\\(pp)} \\
\midrule
TabPFN-TS & MRRC $-$ Base & \centeredtablecell{$-0.48$\\{[$-0.75$, $-0.20$]}} & \centeredtablecell{$-10.51$\\{[$-12.04$, $-9.07$]}} & \centeredtablecell{$-0.29$\\{[$-0.36$, $-0.21$]}} \\
 & MRRC $-$ HFHR & \centeredtablecell{0.02\\{[$-0.10$, 0.13]}} & \centeredtablecell{$-0.62$\\{[$-1.08$, $-0.19$]}} & \centeredtablecell{$-0.64$\\{[$-1.14$, $-0.20$]}} \\
\addlinespace
Chronos-2 & MRRC $-$ Base & \centeredtablecell{$-1.03$\\{[$-1.53$, $-0.39$]}} & \centeredtablecell{$-14.91$\\{[$-16.59$, $-13.35$]}} & \centeredtablecell{$-0.27$\\{[$-0.34$, $-0.19$]}} \\
 & MRRC $-$ HFHR & \centeredtablecell{0.17\\{[$-0.01$, 0.39]}} & \centeredtablecell{0.61\\{[0.16, 1.06]}} & \centeredtablecell{$-0.61$\\{[$-1.11$, $-0.16$]}} \\
\bottomrule
\end{tabular*}
\end{table}